\documentclass{article}

\PassOptionsToPackage{numbers, compress}{natbib}
\PassOptionsToPackage{dvipsnames,x11names}{xcolor}

\usepackage[dandb,final]{neurips_2025}

\usepackage[utf8]{inputenc} 
\usepackage[T1]{fontenc}    
\usepackage{hyperref}       
\usepackage{url}            
\usepackage{booktabs}       
\usepackage{amsfonts}       
\usepackage{nicefrac}       
\usepackage{microtype}      
\usepackage{xcolor}         

\usepackage{tikz}
\usepackage{algorithm}
\usepackage[noend]{algpseudocode}
\usepackage{listings}
\usepackage{graphicx}
\usepackage{caption}
\usepackage{subcaption}
\usepackage{amsthm}
\usepackage{multirow}
\usepackage{makecell}
\usepackage{pgfplots}
\usepackage{etoolbox} 

\usepackage{paralist}

\usepackage[most]{tcolorbox}

\usepackage{pifont}

\newcommand{\cmark}{\textcolor{green!60!black}{\ding{51}}} 
\newcommand{\xmark}{\textcolor{red}{\ding{55}}}             

\usepackage[inline]{enumitem}

\definecolor{lightgray}{gray}{0.95}
\definecolor{darkblue}{rgb}{0.1,0.2,0.6}
\definecolor{codegray}{rgb}{0.4,0.4,0.4}

\lstdefinelanguage{PDDL}{
  morekeywords={define,domain,requirements,types,predicates,action,parameters,precondition,effect,and,not, always, sometime, at-most-once, sometime-before, sometime-after, goal, constraints, exists, forall},
  sensitive=false,
  morecomment=[l]{;},
}

\lstdefinelanguage{nl}{
  sensitive=false,
  morecomment=[l]{;},
}



\lstdefinestyle{pddlstyle}{
  numbers=left,
  numbersep=-6pt,
  backgroundcolor=\color{lightgray},
  basicstyle=\ttfamily\scriptsize,
  keywordstyle=\color{darkblue}\bfseries,
  commentstyle=\color{codegray}\itshape,
  showstringspaces=false,
  frame=single,
  breaklines=true,
  language=PDDL
}

\lstdefinestyle{nl}{
  numbers=left,
  numbersep=-6pt,
  backgroundcolor=\color{lightgray},
  basicstyle=\ttfamily\scriptsize,
  keywordstyle=\color{darkblue}\bfseries,
  commentstyle=\color{codegray}\itshape,
  showstringspaces=false,
  frame=single,
  breaklines=true,
  breakindent=0pt, 
  mathescape=true,
  lineskip=-1pt, 
}

\definecolor{codebg}{rgb}{0.95,0.95,0.95}
\definecolor{keywordcolor}{rgb}{0.26, 0.36, 0.66}
\definecolor{commentcolor}{rgb}{0.25, 0.5, 0.35}
\definecolor{stringcolor}{rgb}{0.58, 0, 0.1}

\lstdefinestyle{mypython}{
    language=Python,
    backgroundcolor=\color{codebg},
    keywordstyle=\color{keywordcolor}\bfseries,
    commentstyle=\color{commentcolor}\itshape,
    stringstyle=\color{stringcolor},
    basicstyle=\ttfamily\footnotesize,
    breaklines=true,
    showstringspaces=false,
    frame=single,
    numbers=left,
    numberstyle=\tiny\color{gray},
    captionpos=b
}

\usetikzlibrary{shapes.misc}
\usetikzlibrary{positioning, arrows.meta, decorations.pathmorphing,shapes}
\everymath{\it}\everydisplay{\it}
\def\val{{=}}

\def\plus{{+}}
\def\minus{{-}}

\usepackage{xspace}

\newenvironment{customlegend}[1][]{%
    \begingroup
    \csname pgfplots@init@cleared@structures\endcsname
    \pgfplotsset{#1}%
}{%
    \csname pgfplots@createlegend\endcsname
    \endgroup
}%
\def\addlegendimage{\csname pgfplots@addlegendimage\endcsname}

\newcommand{\lexicon}{\textsc{LexiCon}\xspace}
\xspaceaddexceptions{'}

\definecolor{lightpurple}{HTML}{C5B4E3}
\definecolor{lightblue}{HTML}{A4DBE8}
\definecolor{lightred}{HTML}{ffb7c5}

\usepackage{bm}
\newcommand{\Exists}{\bm{\exists}\kern-0.6em\bm{\exists}}
\newcommand{\Forall}{\bm{\forall}\kern-0.6em\bm{\forall}}

\usepackage{fontawesome5}
\usetikzlibrary{shapes.geometric,shapes.symbols,arrows,positioning,fit,backgrounds,calc}

\def\append{\mathsf{append}}

\def\pddlProblemUnc{\Pi}
\def\pddlProblem{\Pi^C}
\def\pddlProblemComp{\Pi^{cm}}
\def\pddlAtoms{F}
\def\pddlActions{A}
\def\pddlInit{I}
\def\pddlGoal{G}
\def\pddlConstraints{C}
\def\constr{q}
\def\pddlPreconditions{Pre}
\newcommand{\actionPrec}[1]{\pddlPreconditions(#1)}
\def\pddlEffects{Eff}
\newcommand{\actionEff}[1]{\pddlEffects(#1)}

\def\Always{\mathsf{Always}}
\def\Sometime{\mathsf{Sometime}}
\def\AtMostOnce{\mathsf{AtMostOnce}}
\def\SometimeBefore{\mathsf{SometimeBefore}}
\def\SometimeAfter{\mathsf{SometimeAfter}}

\def\AlwaysShort{\mathsf{A}}
\def\SometimeShort{\mathsf{S}}
\def\AtMostOnceShort{\mathsf{AO}}
\def\SometimeBeforeShort{\mathsf{SB}}
\def\SometimeAfterShort{\mathsf{SA}}

\newcommand{\alw}[1]{\AlwaysShort(#1)}
\newcommand{\st}[1]{\SometimeShort(#1)}
\newcommand{\amo}[1]{\AtMostOnceShort(#1)}
\newcommand{\stb}[2]{\SometimeBeforeShort(#1,#2)}
\newcommand{\sta}[2]{\SometimeAfterShort(#1,#2)}

\def\stateseq{\sigma}
\def\plan{\pi}
\def\planopt{\plan^*}
\def\plannl{\pi_{NL}}
\def\actionnl{a_{NL}}
\def\cost{c}
\def\costopt{\cost^*}
\def\invalid{\mathsf{Invalid}}
\def\suboptimal{\mathsf{Suboptimal}}
\def\optimal{\mathsf{Optimal}}

\newtheoremstyle{mytheorem}
  {3pt}
  {3pt}
  {\normalfont}
  {0pt}
  {\bfseries}
  {.}
  { }
  {}

\theoremstyle{mytheorem}

\newcounter{examplebox}[section]
\renewcommand{\theexamplebox}{\thesection.\arabic{examplebox}}

\newenvironment{examplebox}[1][]{
  \refstepcounter{examplebox}%
  \tcolorbox[
    colframe=blue!70!black,
    colback=white,
    boxrule=0.8pt,
    sharp corners,
    fontupper=\small,
    left=5pt, right=5pt, top=5pt, bottom=5pt,
    title={\textbf{Example \theexamplebox\ifstrempty{#1}{}{:\ #1}}},
    breakable
  ]
}{
  \endtcolorbox
}

\def\pick{\mathtt{pick}}
\def\drop{\mathtt{drop}}

\def\objectInRoom{\mathtt{objectInRoom}}
\def\atObj{\mathtt{at}}
\def\objectColor{\mathtt{objectColor}}
\def\typeof{\mathtt{typeof}}
\def\locked{\mathtt{locked}}
\def\unlocked{\mathtt{unlocked}}

\def\vVar{\mathtt{v}}

\def\d{\mathtt{d}}

\def\ball{\mathtt{ball}}
\def\red{\mathtt{red}}

\def\roomthree{\mathtt{room\_3}}

\def\greendoorone{\mathtt{green\_door\_1}}

\def\redballone{\mathtt{red\_ball\_1}}
\def\redballtwo{\mathtt{red\_ball\_2}}

\def\symk{\mathtt{SymK}}
\def\tcore{\mathtt{TCORE}}
\def\liftedtcore{\mathtt{Lifted TCORE}}

\def\domainformulas{D}
\def\userparams{cfg}
\def\constrop{op}
\def\literalsno{l\_no}

\newcommand{\blockwidth}{0.4cm}
\newcommand{\blockheight}{0.4cm}
\newcommand{\barwidth}{4pt}

\makeatletter
\newcommand\notsotiny{\@setfontsize\notsotiny\@vipt\@viipt}
\makeatother

\title{\lexicon: a Benchmark for Planning\\under Temporal Constraints in Natural Language}

\author{
  Periklis Mantenoglou, Rishi Hazra, Pedro Zuidberg Dos Martires \\
  Örebro University, Sweden \\
  \texttt{\{periklis.mantenoglou, rishi.hazra, pedro.zuidberg-dos-martires\}@oru.se} \\
  \And
  Luc De Raedt \\
  Örebro University, Sweden \& KU Leuven, Belgium \\
  \texttt{luc.deraedt@kuleuven.be} \\
}

\begin{document}

\maketitle

\begin{abstract}
    %
    Owing to their reasoning capabilities, large language models (LLMs) have been evaluated on planning tasks described in natural language.
    However, LLMs have largely been tested on planning domains without constraints.
    In order to deploy them in real-world settings where adherence to constraints, in particular safety constraints, is critical, we need to evaluate their performance on constrained planning tasks.    
    We introduce \lexicon{}---a natural language-based (\textsc{Lexi}) constrained (\textsc{Con}) planning benchmark, consisting of a suite of environments, that can be used to evaluate the planning capabilities of LLMs in a principled fashion.
    The core idea behind \lexicon is to take existing planning environments and impose temporal constraints on the states.
    These constrained problems are then translated into natural language and given to an LLM to solve.
    A key feature of \lexicon is its extensibility. That is, the set of supported environments can be extended with new (unconstrained) environment generators, for which temporal constraints are constructed automatically.
    This renders \lexicon future-proof: the hardness of the generated planning problems can be increased as the planning capabilities of LLMs improve. 
    Our experiments reveal that the performance of state-of-the-art LLMs, including reasoning models like GPT-5, o3, and R1, deteriorates as the degree of constrainedness of the planning tasks increases.
\end{abstract}


\section{Introduction}\label{sec:intro}

Planning with constraints is commonly required in problem-solving settings, ranging from resource allocation and scheduling~\citep{constraints_in_ai_planning} to ensuring safety in reinforcement learning~\citep{altman1995constrained,gattami2021reinforcement,saferl_logic_shields,DBLP:journals/jmlr/GarciaF15}.
%
%
%
Several planning specification languages have been proposed~\cite{DBLP:journals/ai/FikesN71, DBLP:conf/kr/Pednault89, DBLP:journals/aim/McDermott00, DBLP:series/synthesis/2019Haslum}, including formalisms with constraints~\cite{DBLP:journals/ai/GereviniHLSD09}.
However, specifying the complex, possibly compositional, constraints of an environment in a formal language is rather intricate, as it requires, inter alia, significant domain expertise. 
Other common ways of integrating constraints in planning problems is via penalties in a reward function~\cite{aisafetygridworlds, safetygym, carla} or through the physics engine of the environment~\cite{pybullet, mujoco}.
These solutions are also challenging for non-experts, while, after tightly integrating constraints into an environment, they are often difficult to alter if needed.
We address these limitations by enabling the human user to communicate constraints \emph{directly} to the planning agent, via natural language (NL). 
%

The advent of large language models (LLMs), trained on vast textual corpora, has made NL-based planning increasingly feasible. However, whether LLMs possess the reasoning capabilities required for effective planning remains an open question. Some works argue that LLMs can perform reasoning, and even act as \emph{zero-shot planner}~\citep{chain_of_thought,DBLP:conf/nips/KojimaGRMI22,DBLP:conf/icml/HuangAPM22}, while others critically highlight their limitations~\citep{DBLP:conf/nips/ValmeekamMSK23,DBLP:conf/nips/DziriLSLJLWWB0H23,DBLP:conf/nips/StechlyVK24}. In particular, LLM-based planning methods are often inefficient, lack formal guarantees, and incur high computational costs due to the generation of numerous ``thinking tokens''~\citep{DBLP:conf/nips/0001KSS24,DBLP:journals/corr/abs-2410-02162}.

As LLMs are increasingly deployed in domains such as robotics~\citep{saycan,text2motion}, travel planning~\citep{travelplanner}, tool use~\citep{toolformer}, scientific discovery~\citep{aiscientistv2}, and healthcare~\citep{virtuallab}---all of which demand planning and reasoning under constraints---it becomes crucial to rigorously assess their constrained planning capabilities. To this end, we make the following contributions.
\begin{enumerate}[left=0pt,topsep=0pt, partopsep=0pt, itemsep=0pt]
    \item \textbf{Extensible Benchmark.} We introduce \lexicon, an extensible NL-based benchmark for planning with temporal constraints specified on state-trajectories, which is publicly available\footnote{\url{https://github.com/Periklismant/lexicon_neurips}}. It comprises two core components: a symbolic reasoning engine and a translator, which together enable the following functionalities.
    
    \textbf{-- Constrained Problem Generation.}
    This module takes as input an unconstrained planning problem (described in a formal language) and introduces constraints to it, while making sure that it remains solvable. The reasoning engine generates \textbf{\emph{task-aware} constraints} so as to complicate the original problem---resulting in longer solutions compared to its unconstrained version---while guaranteeing that constraints do not subsume one another, which would make them redundant. This leads to a challenging LLM planning benchmark.    
    Crucially, the reasoning engine operates orders of magnitude faster than LLM-based planning, enabling scalable problem generation and evaluation.
    To interface with LLMs, the translator module converts formal planning problems with constraints into NL, leveraging the compositional structure of these problems to produce NL representations in a systematic manner.
    
    \textbf{-- Automated Plan Verification.} The planning capabilities of LLMs are evaluated on the generated NL-representation of constrained planning problems. Subsequently, the reasoning engine automatically verifies whether the LLM-generated plans are correct and/or optimal.
  
    \item \textbf{Experimental Evaluation.}
    We evaluated several state-of-the-art LLMs, including reasoning models like OpenAI o3~\citep{openai-o3}, DeepSeek R1~\citep{deepseek_r1}, Gemini 2.5 Pro~\citep{google2025gemini2.5}, Claude 3.7 Sonnet~\citep{claude_37_sonnet}, and GPT-5~\cite{openai-gpt5}, on benchmarks generated by \lexicon.
     Using constrained problems of increasing compositional complexity, we found that LLM performance consistently declines with the number of constraints, suggesting that current models do not yet match the performance of formal planning algorithms.
    
\end{enumerate}

\begin{figure}[t]
  \centering
    \begin{subfigure}[b]{0.175\textwidth}
    \centering
    \begin{minipage}[t]{\linewidth} 
    \centering
    \resizebox{\linewidth}{!}{\includegraphics[width=\linewidth]{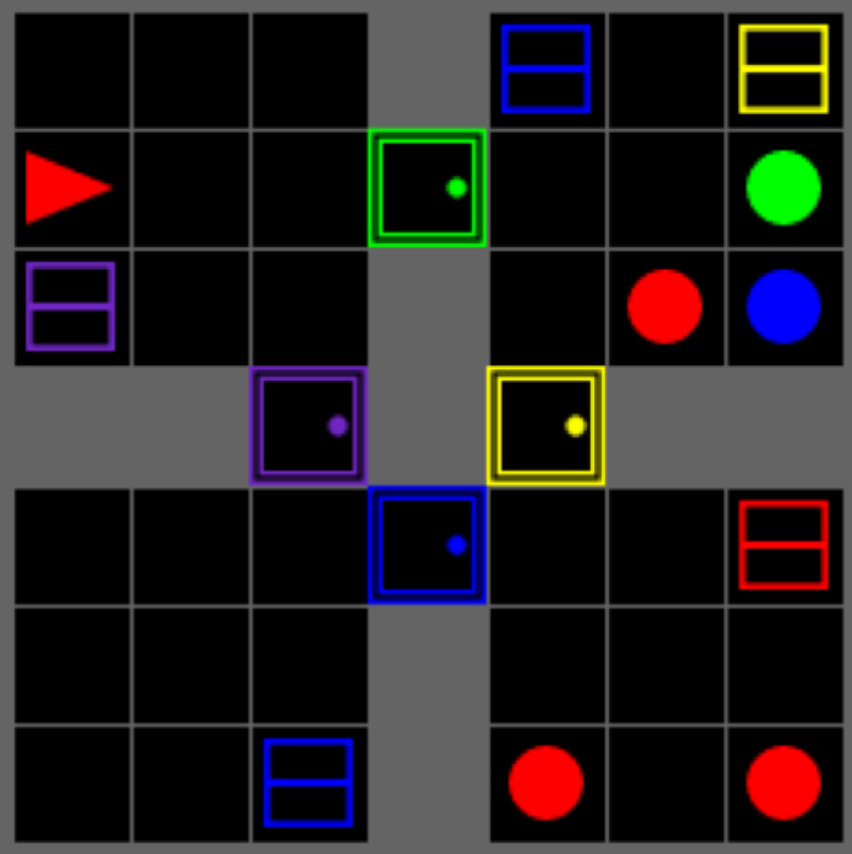}}
    \resizebox{\linewidth}{!}{
        \begin{tcolorbox}[
                colback=lightgray!30,
                colframe=gray!80,
                boxrule=0.5pt,
                arc=2mm,
                left=1mm,
                right=1mm,
                top=0.5mm,
                bottom=0.5mm,
                boxsep=0mm,
                fontupper=\footnotesize,
            ]
                \textbf{Goal}: Reach a red ball. \\
                \textbf{Constraints}: Do not unlock the green door \& get the purple box in the top-right room.
            \end{tcolorbox}
    }
    \end{minipage}
    \end{subfigure}
    \hfill
    \begin{subfigure}[b]{0.2\textwidth}
    \centering
    \begin{minipage}[t]{\linewidth}
    \centering
    \resizebox{\linewidth}{!}{\begin{tikzpicture}[
  block/.style={
    rectangle, draw=black, 
    minimum width=\blockwidth, 
    minimum height=\blockheight, 
    text centered,
    anchor=south
  },
  every node/.style={font=\small},
  arrow/.style={thick, ->, >=Stealth},
]

\coordinate (i1) at (0,0);
\node[block, fill=red!80]   (IA) at ($(i1)+(0,\blockheight)$) {A};
\node[block, fill=blue!70, above=0pt of IA]  (IB) {B};

\coordinate (i2) at (0.6,0);
\node[block, fill=green!70] (IC) at ($(i2)+(0,\blockheight)$) {C};

\coordinate (i3) at (1.2,0);
\node[block, fill=green!70](ID) at ($(i3)+(0,\blockheight)$) {D};
\node[block, fill=red!50, above=0pt of ID]   (IE) {E};
\node[block, fill=blue!40, above=0pt of IE]  (IF) {F};

\coordinate (g1) at (2.4,0);
\node[block, fill=green!70] (GC) at ($(g1)+(0,\blockheight)$) {C};
\node[block, fill=blue!70, above=0pt of GC]  (GB) {B};
\node[block, fill=red!80, above=0pt of GB]   (GA) {A};

\coordinate (g2) at (3,0);
\node[block, fill=green!70](GD) at ($(g2)+(0,\blockheight)$) {D};
\node[block, fill=red!50, above=0pt of GD]   (GE) {E};

\coordinate (g3) at (3.6,0);
\node[block, fill=blue!40]  (GF) at ($(g3)+(0,\blockheight)$) {F};

\node at (0.6, 2.1) {Initial State};
\node at (3, 2.1) {Goal};

\draw[arrow, bend left=25] (0.6,2.3) to (3,2.3);

\end{tikzpicture}}
    \resizebox{\linewidth}{!}{
        \begin{tcolorbox}[
                colback=lightgray!30,
                colframe=gray!80,
                boxrule=0.5pt,
                arc=2mm,
                left=1mm,
                right=1mm,
                top=0.5mm,
                bottom=0.5mm,
                boxsep=0mm,
                fontupper=\footnotesize,
            ]
                \textbf{Goal}: Get block A on top of block B.  \\
                \textbf{Constraints}: Never place block B on the table \& sometime, clear block E.  
            \end{tcolorbox}
    }
    \end{minipage}
    \end{subfigure}
    \hfill
    \begin{subfigure}[b]{0.2\textwidth}
    \centering
    \begin{minipage}[t]{\linewidth}
    \centering
    \resizebox{\linewidth}{!}{\begin{tikzpicture}[
  city/.style={
    draw,
    fill=blue!5,
    decorate,
    decoration={random steps, segment length=4pt, amplitude=2pt},
    minimum width=2cm, minimum height=2cm,
    shape=ellipse
  },
  location/.style={
    circle,
    draw=black,
    fill=gray!20,
    minimum size=6pt,
    inner sep=0pt
  },
  icon/.style={font=\faIcon[regular]{truck}, scale=1.3},
  every node/.style={font=\small}
]

\node[city, label=above:City A] (cityA) at (0,0) {};

\node[location] (a1) at ($(cityA.center)+(-0.3,0.25)$) {};
\node[location] (a2) at ($(cityA.center)+(0.5,0.05)$) {};
\node[location] (a3) at ($(cityA.center)+(-0.2,-0.6)$) {};

\node[font=\faIcon{truck}, above=2pt] at (a1) {};
\node[font=\faIcon{dropbox}, above=2pt] at (a2) {}; 
\node[font=\faIcon{plane}, above=2pt] at (a3) {};

\node[city, label=above:City B] (cityB) at (2.5,0) {};

\node[location] (b1) at ($(cityB.center)+(-0.4,-0.3)$) {};
\node[location] (b2) at ($(cityB.center)+(0.25,0.1)$) {};

\node[font=\faIcon{plane}, above=2pt] at (b1) {};
\node[font=\faIcon{truck}, above=2pt] at (b2) {}; 

\end{tikzpicture}}
    \resizebox{\linewidth}{!}{
        \begin{tcolorbox}[
                colback=lightgray!30,
                colframe=gray!80,
                boxrule=0.5pt,
                arc=2mm,
                left=1mm,
                right=1mm,
                top=0.5mm,
                bottom=0.5mm,
                boxsep=0mm,
                fontupper=\footnotesize,
            ]
                \textbf{Goal}: Move the package to the airport of City B. \\
                \textbf{Constraints}: Never use the airplane that is in City A.  
            \end{tcolorbox}
    }
    \end{minipage}
    \end{subfigure}
    \hfill 
    \begin{subfigure}[b]{0.175\textwidth}
    \centering
    \begin{minipage}[t]{\linewidth} 
    \centering
    \includegraphics[width=\linewidth]{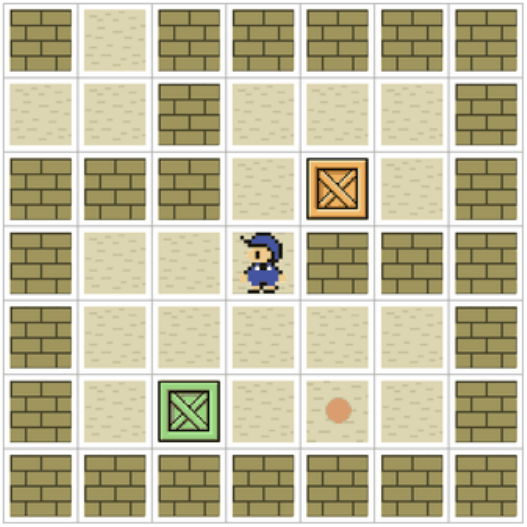} 
    \resizebox{\linewidth}{!}{
        \begin{tcolorbox}[
                colback=lightgray!30,
                colframe=gray!80,
                boxrule=0.5pt,
                arc=2mm,
                left=1mm,
                right=1mm,
                top=0.5mm,
                bottom=0.5mm,
                boxsep=0mm,
                fontupper=\footnotesize,
            ]
                \textbf{Goal}: Move all boxes on goal squares. \\
                \textbf{Constraints}: Sometime, grid square (3,3) must be occupied.
            \end{tcolorbox}
            }
    \end{minipage}
    \end{subfigure}
    \hfill 
    \begin{subfigure}[b]{0.205\textwidth}
    \centering
    \begin{minipage}[t]{\linewidth} 
    \centering
    \resizebox{\linewidth}{!}{\input{figures/tex/alfworld}}
    \resizebox{\linewidth}{!}{
        \begin{tcolorbox}[
                colback=lightgray!30,
                colframe=gray!80,
                boxrule=0.5pt,
                arc=2mm,
                left=1mm,
                right=1mm,
                top=0.5mm,
                bottom=0.5mm,
                boxsep=0mm,
                fontupper=\footnotesize,
            ]
                \textbf{Goal}: Hold a washed apple. \\
                \textbf{Constraints}: Sometime before washing the apple, place a plate on the kitchen counter.
            \end{tcolorbox}
            }
    \end{minipage}
    \end{subfigure}
\caption{Constrained problems on environments supported in \lexicon. From left to right: BabyAI~\citep{babyai}, Blocksworld~\citep{DBLP:journals/ai/GuptaN92}, Logistics~\citep{DBLP:journals/aim/McDermott00}, Sokoban~\citep{DBLP:conf/nips/FengGS20} and AlfWorld~\cite{alfworld}. A constrained planning task is specified by an initial state, a goal, and a set of constraints to be respected.
}
\label{fig:showcase}
\end{figure}

\lexicon\ supports five environments (Figure~\ref{fig:showcase}) and is designed to be extensible. We expect it to remain valuable even as more capable LLMs emerge. As LLMs improve, \lexicon\ can adapt by generating problems with increased constraint complexity or encorporating new environments, resulting in a flexible, future-proof benchmark that does not rely on static planning problems. Unless LLMs truly acquire algorithmic planning abilities---generalizing across problem instances like symbolic planners---we expect \lexicon\ to continue serving as an effective tool for assessing their planning capabilities.

\section{Related Work}\label{sec:related}

\begin{table}[t]
\centering
\begin{tabular}{lccccc}
\textbf{Benchmark} & 
\makecell[c]{Constraints} & 
\makecell[c]{NL\\Interface} & 
\makecell[c]{Automated\\Curation} & 
\makecell[c]{Suite\\Extensibility} & 
\makecell[c]{Environment\\Diversity} \\
\hline 
\\
BabyAI~\citep{babyai}         & 
\xmark & 
\cmark &
\cmark & 
\cmark & 
\xmark \\
AlfWorld~\citep{alfworld}       & 
\xmark & 
\cmark & 
\cmark & 
\xmark & 
\xmark \\
PlanBench~\citep{planbench}      & 
\xmark & 
\cmark & 
\cmark & 
\cmark & 
\cmark \\
ACPBench~\citep{acpbench}      & 
\xmark & 
\cmark & 
\cmark & 
\cmark & 
\cmark \\
BALROG~\citep{balrog}         & 
\xmark & 
\cmark & 
\cmark & 
\cmark & 
\cmark \\
Safety Gym~\citep{safetygym}     & 
\cmark & 
\xmark & 
\cmark & 
\xmark & 
\xmark \\
TravelPlanner~\citep{travelplanner}  & 
\cmark & 
\cmark & 
\xmark & 
\xmark & 
\xmark \\
Natural Plan~\citep{naturalplan}   & 
\cmark & 
\cmark & 
\xmark & 
\xmark & 
\cmark \\
\textbf{\lexicon} & 
\cmark & 
\cmark & 
\cmark & 
\cmark & 
\cmark \\
\rule{0pt}{2.5ex}
\end{tabular}
\caption{Comparison of simulation benchmarks. ``Automated curation'' indicates the ability to automatically generate new planning problem instances and verify solutions for those instances. ``Suite Extensibility'' requires that new planning domains can be added to the benchmark without rewriting its code. ``Environment Diversity'' indicates that the benchmark supports more than one type of planning domain (e.g., it is not restricted solely to 2D gridworld problems).}
\label{tab:benchmark-comparison}
\end{table}

Table \ref{tab:benchmark-comparison} compares \lexicon\ with state-of-the-art planning benchmarks.
Benchmarking the planning capabilities of LLMs requires an NL interface, which limits the applicability of traditional constrained environments such as Safety Gym~\citep{safetygym} that lack NL support. While simulators such as BabyAI~\citep{babyai}, gComm~\citep{gcomm}, and AlfWorld~\citep{alfworld} support NL interaction, they do not model constraints and are limited to narrow domains (e.g., 2D grids or household settings). Constrained planning benchmarks like NaturalPlan~\citep{naturalplan} and TravelPlanner~\citep{travelplanner} also support NL, but their tasks are either manually curated or carefully constructed offline, resulting in limited extensibility. Additionally, verifying LLM-generated plans in these settings typically requires exhaustively enumerating all valid solutions, which is prohibitively expensive. 
In contrast, \lexicon\ supports the generation of a potentially unbounded number of constrained tasks and can automatically verify agent outputs using its reasoning engine. This enables rigorous, scalable evaluation without needing exhaustive (manual) plan enumeration. 

While one might consider augmenting planning benchmarks such as PlanBench~\citep{planbench} or BALROG~\citep{balrog} with constraint-handling functionalities, these systems lack the infrastructure to synthesize, solve, and validate constrained tasks in an integrated manner. In contrast, \lexicon\ was built from the ground up to support automated constraint generation, enforcement, and verification. As LLMs continue to improve in their reasoning capabilities~\citep{have-llms-learned-to-reason}, \lexicon\ provides a principled platform for evaluating them on increasingly complex planning tasks with compositional constraints.
Moreover, its reasoning engine is domain-agnostic, facilitating seamless extension to new environments (i.e., suite extensibility). In what follows, we illustrate the planning formalism in \lexicon, and describe its architecture.

\section{The \lexicon Simulator}\label{sec:lexicon}


\subsection{Planning Specification Language}\label{sec:language}

\lexicon\ supports planning problems expressed in PDDL3.0, an extension of the PDDL formal planning language that includes constraints~\cite{DBLP:journals/ai/GereviniHLSD09}.

\begin{figure}[t]
  \centering
  \begin{minipage}[c]{0.27\textwidth}
    \centering
    \includegraphics[width=\linewidth]{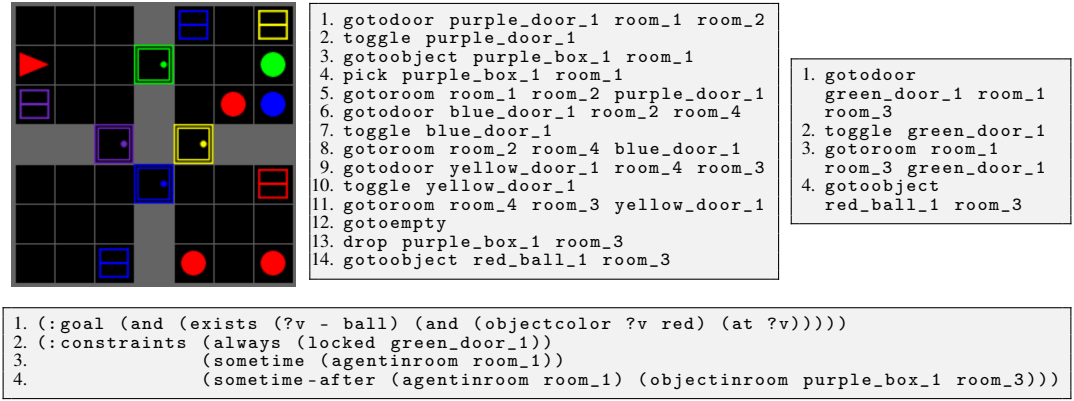} 
  \end{minipage}%
  \hfill
  \begin{minipage}[c]{0.43\linewidth}
    \lstset{style=nl} 
    \begin{lstlisting}
  gotodoor purple_door_1 room_1 room_2
  toggle purple_door_1
  gotoobject purple_box_1 room_1
  pick purple_box_1 room_1
  gotoroom room_1 room_2 purple_door_1
  gotodoor blue_door_1 room_2 room_4
  toggle blue_door_1
  gotoroom room_2 room_4 blue_door_1
  gotodoor yellow_door_1 room_4 room_3
  toggle yellow_door_1
  gotoroom room_4 room_3 yellow_door_1
  gotoempty
  drop purple_box_1 room_3
  gotoobject red_ball_1 room_3
    \end{lstlisting}
  \end{minipage}
  \hfill
  \begin{minipage}[c]{0.25\textwidth}
    \lstset{style=nl} 
    \begin{lstlisting}
  gotodoor green_door_1 room_1 room_3
  toggle green_door_1
  gotoroom room_1 room_3 green_door_1
  gotoobject red_ball_1 room_3
    \end{lstlisting}
  \end{minipage}
  \vfill
  \begin{minipage}[c]{\textwidth}
    \lstset{style=nl} 
    \begin{lstlisting}
  (:goal (and (exists (?v - ball) (and (objectcolor ?v red) (at ?v)))))
  (:constraints (always (locked green_door_1))
                (sometime (agentinroom room_1))
                (sometime-after (agentinroom room_1) (objectinroom purple_box_1 room_3)))
    \end{lstlisting}
  \end{minipage}  
\caption{Left: The initial state of the constrained planning problem in Example \ref{ex:running}. The red triangle represents the agent. Bottom: The goal and the constraints of the problem in PDDL3.0. Middle: Optimal plan for this problem. Right: Optimal plan for the corresponding unconstrained problem.}\label{fig:running}
\end{figure}

\begin{examplebox}[Constrained Planning in BabyAI]\label{ex:running}
BabyAI contains problems where an agent needs to navigate the rooms of 2D gridworld, while interacting with objects, to complete some task~\cite{babyai}.
Figure \ref{fig:running} (left) shows the initial state of a problem from BabyAI.
%
%
%
BabyAI problems are grounded in PDDL; a domain file specifies the object types, the (time-varying) state atoms, and the actions of the domain, while a problem file denotes the objects of the puzzle, the initial state, the goal, and the constraints.
In this case, the domain file defines atom $\locked(\d)$, expressing that door $\d$ is locked, and action $\pick$, outlining the conditions for and the effects of picking up and holding an object.
%
%
%
%
%
%
Figure \ref{fig:running} (bottom) outlines the goal and the constraints of the problem.
The goal is to reach a red ball, while the constraints dictate that (i) the agent must never unlock the green door, (ii) at some point, the agent must visit room 1 (top-left room), and (iii) some time after visiting room 1, purple box 1 needs to be in room 3 (top-right room).
%
%
\end{examplebox}

In \lexicon, we are interested in \textbf{optimal planning}, i.e., finding a plan that (1) reaches the goal while satisfying all constraints and (2) has minimum length.
Optimal planning on the problem described in Example \ref{ex:running} is easy if the constraints are ignored---an optimal plan for the unconstrained problem consists of 4 actions (see Figure \ref{fig:running} (right)). However, the constrained version is significantly more challenging. For example, to satisfy the constraint that the green door must always remain locked, the agent must take a longer path through the purple and blue doors to reach the room containing the red ball---resulting in 14 actions (cf., Figure~\ref{fig:running} (middle)).
%
%
%
%

\subsection{The \lexicon\ Architecture}\label{sec:architecture}

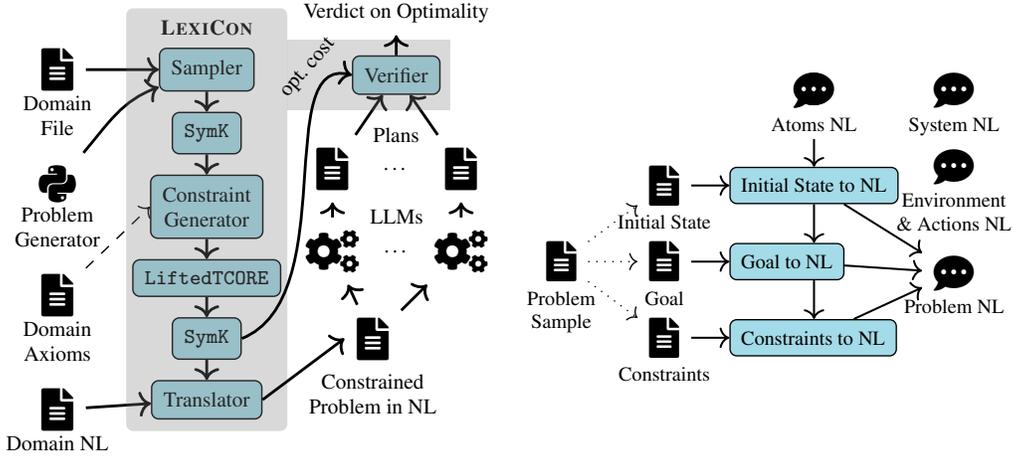
\begin{figure}[htbp]
\begin{minipage}[c]{.49\linewidth}
\resizebox{\linewidth}{!}{\begin{tikzpicture}[
      every  node/.style={font=\tiny},
      process/.style={rectangle, fill=lightblue, draw, text centered, minimum width=0.01cm, minimum height=0.01cm, rounded corners=2pt},
      archbackground/.style={fill=black!50, opacity=0.3, minimum width=1.6cm, minimum height=4.2cm, rounded corners=2pt},
      llmsbackground/.style={fill=black!50, opacity=0.3, minimum width=2.5cm, minimum height=0.8cm, rounded corners=2pt},
      verifierbackground/.style={fill=black!50, opacity=0.3, minimum width=1.62cm, minimum height=0.7cm},
      datastore/.style={draw, fill=lightpurple, rounded rectangle, rounded rectangle east arc=concave, rounded rectangle arc length=150},
      decision/.style={diamond, aspect=0.65, scale=0.9, fill=lightpurple, draw, text centered, minimum width=0.01cm, minimum height=0.01cm},
      note/.style={ellipse, draw, text width=1cm, text centered, minimum height=2cm},
      subproc/.style={process, fill=lightred, dashed, minimum width=0.5cm, minimum height=0.5cm},
      pyicon/.style={draw=none, minimum width=1cm, minimum height=1cm, path picture={
          \node at (path picture bounding box.center) {\large\faIcon{python}};
      }},
      planicon/.style={rectangle, draw, minimum width=3cm, minimum height=2cm, path picture={
          \foreach \y in {0.3,0.6,...,1.2} {
              \draw[thick] ([yshift=-\y cm]path picture bounding box.north west) -- ([yshift=-\y cm]path picture bounding box.north east);
          }
      }},
      node distance=2cm
  ]

  
  \node (domain) at (0, 0) {\large\faIcon{file-alt}};
  \node[below=-0.2cm of domain, align=center] {Domain\\ File};

  \node (generator) at ($(domain.south) + (0, -0.8cm)$) {\large\faIcon{python}};
  \node[below=-0.2cm of generator, align=center] {Problem\\ Generator};

  \node (formulas) at ($(generator.south) + (0, -0.8cm)$){\large\faIcon{file-alt}};
  \node[below=-0.2cm of formulas, align=center] {Domain\\ Axioms};

  \node (domainnl) at ($(formulas.south) + (0, -0.8cm)$){\large\faIcon{file-alt}};
  \node[below=-0.2cm of domainnl, align=center] {Domain NL};
  

  \node[process, align=center] at ($(domain.east) + (1.2cm, 0)$) (stage1) {Sampler};
  
  \draw[->, thick] (domain) -- (stage1);
  \draw[->, thick] (generator) to[out=50, in=200]  (stage1);
   
  \node[process, anchor=north] at ($(stage1.south) + (0, -0.2cm)$) (stage2) {$\symk$};
  
  \node[process, align=center, anchor=north] at ($(stage2.south) + (0, -0.2cm)$) (stage3) {Constraint\\ Generator};
 i 
  \draw[->, dashed] (formulas) -- (stage3.west);
  
  \node[process, align=center, anchor=north] at ($(stage3.south) + (0, -0.2cm)$) (stage4) {$\liftedtcore$};
  
  \node[process, anchor=north] at ($(stage4.south) + (0, -0.2cm)$) (stage5) {$\symk$};

  \node[process, align=center, anchor=north] (stage6) at ($(stage5.south) + (0, -0.2cm)$) {Translator};

  \draw[->, thick] (domainnl) -- (stage6);
  
  \draw[->, thick] (stage1.south) -- (stage2.north);
  \draw[->, thick] (stage2.south) -- (stage3.north);
  \draw[->, thick] (stage3.south) -- (stage4.north);
  \draw[->, thick] (stage4.south) -- (stage5.north);
  \draw[->, thick] (stage5.south) -- (stage6.north);

  \node at ($(stage1.north) + (0, 0.2cm)$) (lexiconlabel) {\textbf{\lexicon}};
  
  \node[archbackground, anchor=north] at (lexiconlabel.north) (base) {};

  
  \node (problem) at ($(stage5.east) + (1.3cm, 0)$) {\large\faIcon{file-alt}};
  \node[below=-0.1cm of problem, align=center] (problemlabel) {Constrained\\ Problem in NL};
  \draw[->, thick] (stage6.east) -- (problem.west);


  \node[anchor=south] (llm1) at ($(problem.north) + (-0.4cm, 0.2cm)$) {\large\faIcon{cogs}};
  
  \node[anchor=west] (llmdots) at (llm1.east) {\dots};
  
  \node[anchor=west] (llmn) at (llmdots.east) {\large\faIcon{cogs}};
  
  \node[] at ($(llmdots.north) + (0, 0.2cm)$) (llmslabel) {LLMs};

  \draw[->, thick] (problem) -- (llm1);
  \draw[->, thick] (problem) -- (llmn);
  
  \node[anchor=south] at ($(llm1.north) + (0, 0.15cm)$) (plan1) {\large\faIcon{file-alt}};
  
  \node[anchor=west] at ($(plan1.east) + (0.1cm, 0)$) (plandots) {\dots};
  
  \node[anchor=south] at ($(llmn.north) + (0, 0.15cm)$) (plann) {\large\faIcon{file-alt}};
  
  \node[] at ($(plandots.north) + (0, 0.2cm)$) (planslabel) {Plans};
  
  \draw[->, thick] (llm1) -- (plan1);
  \draw[->, thick] (llmn) -- (plann);

  \node[process, align=center, anchor=south] at ($(planslabel.north) + (0, +0.2cm)$) (verifier) {Verifier};
    
  \draw[->, thick] (plan1) -- (verifier);
  \draw[->, thick] (plann) -- (verifier);

  \node[verifierbackground, anchor=east] at ($(verifier.east) + (0.1cm, 0)$) (verifierbase) {};
  
  \node[anchor=south] (verdict) at ($(verifier.north)+(0, 0.2cm)$) {Verdict on Optimality};
  \draw[->, thick] (verifier) -- (verdict);
  \draw[->, thick] (stage5.0) .. controls +(1,0.4) and +(-0.8,0.2) .. node[pos=0.85, sloped, above] {opt.~cost} (verifier.180);

  
  \end{tikzpicture}}
\end{minipage}
\begin{minipage}[c]{.49\linewidth}
    \resizebox{\linewidth}{!}{\begin{tikzpicture}[
      every  node/.style={font=\tiny},
      process/.style={rectangle, fill=lightblue, draw, text centered, minimum width=0.7cm, minimum height=0.2cm, rounded corners=2pt},
      datastore/.style={draw, fill=lightpurple, rounded rectangle, rounded rectangle east arc=concave, rounded rectangle arc length=150},
      decision/.style={diamond, aspect=0.65, scale=0.9, fill=lightpurple, draw, text centered, minimum width=0.01cm, minimum height=0.01cm},
      extract/.style={dotted, ->},
      ioarrow/.style={->, semithick},
      note/.style={ellipse, draw, text width=1cm, text centered, minimum height=2cm},
      subproc/.style={process, fill=lightred, dashed, minimum width=0.5cm, minimum height=0.5cm},
      pyicon/.style={draw=none, minimum width=1cm, minimum height=1cm, path picture={
          \node at (path picture bounding box.center) {\large\faIcon{python}};
      }},
      planicon/.style={rectangle, draw, minimum width=3cm, minimum height=2cm, path picture={
          \foreach \y in {0.3,0.6,...,1.2} {
              \draw[thick] ([yshift=-\y cm]path picture bounding box.north west) -- ([yshift=-\y cm]path picture bounding box.north east);
          }
      }},
      node distance=2cm
  ]

    \node (problem) at (0,0) {\large\faIcon{file-alt}};
    \node[below=-0.15cm of problem, align=center] (problemlabel) {Problem\\ Sample};

  
    \node (initstate) at ($(problem.east) + (0.8cm, 0.8cm)$) {\large\faIcon{file-alt}};
    
    \draw[extract] (problem) -- (initstate);
    \node[below=-0.15cm of initstate, align=center] {Initial State};
  
    \node[process, right=0.4cm of initstate, align=center] (init2nl) {Initial State to NL};
    \draw[ioarrow] (initstate) -- (init2nl);
  
  
    \node (goal) at ($(problem.east) + (0.8cm, 0cm)$) {\large\faIcon{file-alt}};
    
    \draw[extract] (problem) --  (goal);
    \node[below=-0.15cm of goal, align=center] {Goal};
  
    \node[process, right=0.4cm of goal, align=center] (goal2nl) {Goal to NL};
    \draw[ioarrow] (goal) -- (goal2nl);
  
  
    \node (constraints) at ($(problem.east) + (0.8cm, -0.8cm)$) {\large\faIcon{file-alt}};
    
    \draw[extract] (problem) -- (constraints);
    \node[below=-0.15cm of constraints, align=center] {Constraints};
  
    \node[process, right=0.4cm of constraints, align=center] (constraints2nl) {Constraints to NL};
    \draw[ioarrow] (constraints) -- (constraints2nl);
  
  
    
  
  
    \node (fluentdescr) at ($(init2nl.north) + (0, 0.8cm)$) {\large\faIcon{comment-dots}} ;
    \node[below=-0.15cm of fluentdescr, align=center] (fluentdescrlabel) {Atoms NL};
    
    \draw[ioarrow] ($(fluentdescr.south) + (0, -0.2cm)$)  -- (init2nl);

    \draw[ioarrow] (init2nl.south) --  (init2nl|-goal2nl.north);

    \draw[ioarrow] (init2nl|-goal2nl.south) --  (init2nl|-constraints2nl.north);
    
    \node (fluentdescr) at ($(init2nl.north) + (0, 0.8cm)$) {\large\faIcon{comment-dots}} ;
    
    \node[right=0.8cm of fluentdescr] (systemprompt) {\large\faIcon{comment-dots}} ;
    \node[below=-0.15cm of systemprompt, align=center] (systempromptlabel) {System NL};
    
    \node[below=0.18cm of systemprompt] (domaindescr) {\large\faIcon{comment-dots}} ;
    \node[below=-0.15cm of domaindescr, align=center] (domaindescrlabel) {Environment\\ \& Actions NL};
    
    \node[below=0.5cm of domaindescr] (problemdescr) {\large\faIcon{comment-dots}} ;
    \node[below=-0.15cm of problemdescr, align=center] (problemdescrlabel) {Problem NL};
  
    \draw[ioarrow] (init2nl) to (problemdescr);
    \draw[ioarrow] (goal2nl) to (problemdescr);
    \draw[ioarrow] (constraints2nl) to (problemdescr);
  \end{tikzpicture}} 
\end{minipage}
\caption{Left: The architecture of \lexicon. Solid arrows denote input/output data transfers. Dashed arrows denote optional input. Right: The translator of \lexicon. Dotted arrows express content extraction.}
\label{fig:lexicon}
\end{figure}

Figure \ref{fig:lexicon}(left) illustrates the architecture of \lexicon.
The modules between the ``Sampler'' and the ``Translator'' implement the constrained problem generator functionality of \lexicon, while the ``Verifier'' module realises the automated plan verification functionality. 
We first outline constrained problem generation in PDDL, then its translation into natural language, and lastly our plan verifier.

\textbf{Constrained Planning Problem Generator.} We developed a constrained PDDL problem generator, extending the literature with a \textbf{task-aware} method for producing constraints for arbitrary PDDL problems.
Its task is to generate constrained planning problems along with their optimal cost.
The generator first samples an unconstrained problem using a domain file and a (unconstrained) state-goal pair generator, and then computes an optimal plan for the problem using the state-of-the-art planner $\symk$~\cite{DBLP:journals/ai/TorralbaAKE17}.
This plan, along with the unconstrained problem, is passed to \lexicon's constraint generator, which synthesizes task-aware constraints that (1) preserve feasibility, i.e., the problem still has a solution, and (2) increase the optimal cost relative to the unconstrained version. 
%
%
%
%

\begin{examplebox}[Constraint Generation in BabyAI]\label{ex:literal_sampling}
Consider the unconstrained plan in Figure~\ref{fig:running}. 
%
%
Using \lexicon, we can automatically construct an $\Always(\phi)$ constraint by analyzing the state transitions induced by this plan. 
The system samples domain atoms and evaluates their suitability for inclusion in $\phi$ based on problem complication, consistency, and non-redundancy.
%
%
%
For example, atom $\atObj(\redballtwo)$ is excluded since it does not hold in the initial state and thus cannot ``always'' hold.
%
Similarly, $\objectInRoom(\redballone, \roomthree)$ is not selected as it holds in all states of the unconstrained plan, thus offering no added difficulty.
%
In contrast, $\locked(\greendoorone)$ is included in $\phi$, as enforcing it prevents use of the green door---forcing a detour through the purple and blue doors---which increases the plan's optimal cost (see Figure~\ref{fig:lexicon} (left)).
%


%
%
\end{examplebox}

In \lexicon, users can optionally provide atemporal \textbf{domain axioms} to guide the constraint generator toward meaningful, non-conflicting constraints. For example, given the axiom $\forall d{:}\ \neg (\locked(d) \wedge \unlocked(d))$ and an existing constraint $\Always(\locked(\greendoorone))$, the generator avoids sampling $\Sometime(\unlocked(\greendoorone))$ since it would be unsatisfiable.

Next, we compute the optimal cost of the constrained problem generated by \lexicon, which is necessary to evaluate LLM outputs against ground-truth optimal plans. However, no existing planner supports constrained planning problems with actions that have conditional effects, which are often essential to specify certain domains, such as BabyAI. To overcome this, we compile the constraints away~\citep{DBLP:conf/kr/WrightMN18,DBLP:conf/aips/PercassiG19,DBLP:conf/aaai/BonassiGS24}, producing an equivalent problem without constraints, which can be solved by $\symk$. \lexicon\ uses the $\tcore$ compiler~\citep{DBLP:conf/aips/BonassiGPS21} for this translation. To avoid the cost of grounding, we apply a lifted variant of $\tcore$. Solving the compiled problem with $\symk$ yields a formally verified optimal cost for the original constrained planning problem.

Our constrained problem generator is \textbf{compositional}, allowing users to control the number and complexity of constraints, enabling the generation of increasingly challenging benchmarks for future LLMs. It is also \textbf{extensible}: to support a new domain, users need only provide a PDDL domain file and an automated initial state–goal generator, avoiding manual problem construction. Domains can also be specified in Python via the Unified Planning framework~\cite{DBLP:journals/softx/MicheliBRSVFRTBGIIKPSSS25}, easing use for non-experts.


%
%

\textbf{PDDL to Natural Language Translator.} To evaluate LLMs on constrained problems, we first translate them into natural language (NL). As shown in Figure~\ref{fig:lexicon} (right), our translator extracts the instance-specific elements---initial state, goal, and constraints---and composes a problem prompt in NL. Since these instance-specific elements are built compositionally from domain atoms, their NL descriptions are generated by combining predefined NL templates for each atom.
Domain-level descriptions (e.g., environment and action semantics) are carefully handcrafted per domain.

\begin{figure}[t]
    \lstset{style=nl} 
    \begin{lstlisting}
  The original state of the world is:
   `you are in room_1'
   `purple_box_1 is in room_1'
   `blue_box_1 is in room_2'
   <Description of the remaining atoms that hold initially>

  The task is to bring about the following situation:
   `There is a ball v such that `The following conditions are all true: `v is red', `you are in front of v'''

  A valid plan for the abovementioned problem must abide by the following constraints:
   `The following expression must hold in every state: `green_door_1 is locked''
   `The following expression must hold in at least one state: `you are in room_1''
   `If expression `you are in room_1' holds in some state s, then expression `purple_box_1 is in room_3' must hold at s or at some state after s'
    \end{lstlisting}
    \lstset{style=nl} 
    \begin{lstlisting}
  Provided a planning problem, consisting of an initial state of the world, a final goal and a set of constraints, your task is to provide a valid sequence of actions that solves the planning problem, i.e., bringing about the goal of the problem while satisfying all constraints.
  You need to provide an optimal plan, i.e., a valid plan whose length is equal or less than the length of any other valid plan.
    \end{lstlisting}
\caption{Top: Fragment of our natural language description of the constrained problem of Example \ref{ex:running}. Bottom: System role prompt.}\label{fig:problem_nl}
\end{figure} 

\begin{examplebox}[NL Translation for BabyAI Problem]
Consider the constrained planning problem in Example~\ref{ex:running}. Figure~\ref{fig:problem_nl} (top) shows a fragment of the NL description generated for this problem by our translator. Lines 2--5 describe the initial state by listing NL descriptions of atoms that hold initially. The goal---``\emph{reach a red ball}''---is represented by the logical formula $\exists \vVar{:}\ \typeof(\vVar,\ball) \wedge \objectColor(\vVar, \red) \wedge \atObj(\vVar)$, which we translate recursively into NL (lines 7--8). This involves mapping the quantifier to ``\emph{There is a ball v such that}'', followed by ``\emph{The following conditions are all true}'', and then enumerating atom-level descriptions. Constraints are translated similarly using this recursive procedure (see lines 10--13).
    %
    %
\end{examplebox}

Our translator is also \textbf{extensible}: to support a new planning domain, one only needs to provide (i) an NL description of the environment and actions, and (ii) NL descriptions for each atom. This eliminates the need for instance-specific NL annotations, allowing the translator to operate directly on any generated constrained problem within the domain.
%
%

\textbf{Automated LLM Plan Verifier.} With \lexicon's modules for generating constrained planning problems in NL in place, we now evaluate LLMs on these problems. Each LLM is given the NL description of a problem along with a fixed system role prompt (Figure~\ref{fig:problem_nl} (bottom)), instructing it to act as an optimal planner. This prompt is used consistently across all domains.

To assess LLM outputs, \lexicon includes a verifier module (Figure~\ref{fig:lexicon} (left)) with three steps:
(1) LLM-generated plans are mapped to PDDL actions using the prescribed output format; deviations are corrected by matching the LLM action to the closest domain action, according to the edit distance~\cite{DBLP:journals/csur/Navarro01}.
(2) The plan is validated using an automated plan validator on the compiled version of the constrained problem produced by $\liftedtcore$, leveraging the guarantee that a plan valid for the compiled problem also satisfies the original constrained problem~\cite{DBLP:conf/aips/BonassiGPS21}.
(3) If valid, the plan is checked for optimality by comparing its length to the optimal cost, which was computed at the problem generation phase.

A rigorous formulation of the constrained planning problem (i.e., with temporal constraints) along with how constrained plans are generated and verified through our reasoning engine is provided in Appendix~\ref{sec:planning problem}.

\begin{figure}[t]
  \begin{minipage}[t]{0.32\textwidth}
    \lstset{style=nl} 
    \begin{lstlisting}
  pick purple_box_1 room_1
  gotodoor purple_door_1 room_1 room_2
  toggle purple_door_1
  gotoroom room_1 room_2 purple_door_1
  gotodoor blue_door_1 room_2 room_4
  toggle blue_door_1
  gotoroom room_2 room_4 blue_door_1
  gotodoor yellow_door_1 room_4 room_3
  toggle yellow_door_1
  gotoroom room_4 room_3 yellow_door_1
  drop purple_box_1 room_3
  gotoobject red_ball_1 room_3 
    \end{lstlisting}
  \end{minipage}
  \hfill
  \begin{minipage}[t]{0.32\linewidth}
    \lstset{style=nl} 
    \begin{lstlisting}
  gotoobject purple_box_1 room_1
  pick purple_box_1 room_1
  gotodoor purple_door_1 room_1 room_2
  toggle purple_door_1
  gotoroom room_1 room_2 purple_door_1
  gotodoor blue_door_1 room_2 room_4
  drop purple_box_1 room_2
  toggle blue_door_1
  gotoroom room_2 room_4 blue_door_1
  gotoobject red_ball_2 room_4
\end{lstlisting}
  \end{minipage}
    \hfill
   \begin{minipage}[t]{0.32\linewidth}
    \lstset{style=nl} 
    \begin{lstlisting}
  gotodoor purple_door_1 room_1 room_2
  toggle purple_door_1
  gotoroom room_1 room_2 purple_door_1
  gotodoor blue_door_1 room_2 room_4
  toggle blue_door_1
  gotoroom room_2 room_4 blue_door_1
  gotoobject red_ball_2 room_4
    \end{lstlisting}
  \end{minipage} 
\caption{Invalid plans suggested by LLMs for the constrained problem in Example \ref{ex:running}.}\label{fig:llms_on_running_example}
\end{figure}

Figure \ref{fig:llms_on_running_example} displays LLM-generated plans for the constrained problem in Example \ref{ex:running}.
The plan on the left was generated by o3; this plan is invalid because it violates the preconditions of the $\pick$ action, i.e., the agent attempts to pick up a purple box at a time when it is not facing that box (cf.~line 1 of Figure \ref{fig:llms_on_running_example} (left) and the starting state in Figure \ref{fig:running} (left)).
The plan in the middle was suggested by Claude 3.7 Sonnet with extended thinking. 
This plan is invalid because the agent attempts to drop an object at a time when it is facing a door instead of an empty position, as required by the preconditions of the $\drop$ action (cf.~lines 6 and 7 in Figure \ref{fig:llms_on_running_example} (middle)).
This type of error may be due to LLM state hallucination or loss of state tracking.
The plan on the right was produced by R1.
This plan ignores the constraint stipulating that the purple box must be placed in the top-right room, and is thus invalid.
Next, we present a thorough evaluation of LLMs on benchmarks generated by \lexicon.

\section{LLM Evaluation on \lexicon}\label{sec:experiments}

\subsection{Evaluation Setup}
\label{subsection: evaluation setup}

Figure \ref{fig:showcase} displays the domains supported in \lexicon, which are:
%
%
%
\begin{compactitem}
\item \textbf{BabyAI}~\citep{babyai}: an environment with minigrid problems, like our running example.
\item \textbf{Blocksworld}: a puzzle where an agent rearranges blocks into a target configuration. Constraints may forbid placing certain blocks on the table or require specific sequences of block manipulations.
\item \textbf{Logistics}: a world consisting of several locations, possibly including packages, trucks and airplanes, where the task is to move all packages to their designated destinations.
Constraints may, e.g., forbid the usage of a specific truck or an airport.
\item \textbf{Sokoban}: a gridworld where an agent has to move a collection of boxes onto target locations. 
Constraints may indicate, e.g., that a grid square must be occupied or cleared.
\item \textbf{AlfWorld}: an environment for executing household task, like putting a book in a drawer, washing and slicing an apple, or turning on a lamp.
Constraints may, e.g., prohibit the use of certain utensils, or impose a (partial) ordering among sub-tasks.
\end{compactitem}

LLMs were tasked with optimal planning on constrained problems generated by \lexicon. Our experiments ran on a standard PC (Ubuntu 22, Ryzen 7 5700U, 16GB RAM), using each LLM's official API and the maximum allowed token limits for completions and reasoning.
The LLM execution parameters for all our experiments are provided in Appendix~\ref{appendix: reproducibility}.

\subsection{Evaluation Results}
\label{subsection: evaluation results}

\begin{figure}[t]
    \centering
    \resizebox{.45\linewidth}{!}{
      \begin{tikzpicture}
    \begin{axis}[
        width=8cm,height=4cm,
        ymin = -1,
        ymax = 34,
        legend style={
          at={(1,1)},
          anchor=north east,
        },
        title={\textbf{Blocksworld}},
        title style={yshift=-5pt},
        xticklabel style={yshift=3pt, font=\footnotesize},
        ylabel style={yshift=-8pt},
        xlabel = {Constraints No (Avg.~Optimal Cost)},
        xlabel style={align=center, font=\footnotesize\bf, yshift=4pt},   
        ytick = {0, 15, 30},
        yticklabels = {0, 0.5, 1}, 
        yticklabel style={text height=5pt, font=\footnotesize},
        ylabel = {Performance},
        ylabel style={align=center, font=\footnotesize\bf, yshift=-4pt}, 
        symbolic x coords = {x0, x1, x3, x5, x7, x10},
        xticklabels = {0(2), 1(6), 3(9), 5(12), 7(14), 10(16)},
        xtick={x0, x1, x3, x5, x7, x10}, 
        xtick style={draw=none},
        ybar=0pt,
        bar width=\barwidth, 
        xtick distance=1,               
        enlarge x limits=0.1, 
      ]

    \def\CustomLabelsR{{10, 9, 19, 14, 30, }} 
    \addplot[forget plot,
      nodes near coords style={yshift=-2pt},
      fill=red!15!white, draw=none, bar shift=-2.5*\barwidth]
      coordinates {(x0,26) (x1,24) (x3,15) (x5,9) (x7,5) (x10,0)};

    \def\CustomLabelsO{{1, 9, 14, 10, 18, 35}} 
    \addplot[forget plot,
        nodes near coords style={yshift=-2pt},
        fill=blue!15!white, draw=none, bar shift=-1.5*\barwidth]
        coordinates {(x0,30) (x1,28) (x3,20) (x5,19) (x7,14) (x10,5)};
      
    \def\CustomLabelsG{{ , 8, 22, 24, 22, 25}} 
    \addplot[forget plot,
    nodes near coords style={yshift=-2pt},
    fill=green!15!white, draw=none, bar shift=-.5*\barwidth]
      coordinates {(x0,30) (x1,29) (x3,23) (x5,18) (x7,8) (x10,4)};

    \def\CustomLabelsC{{13, 30, 25, 33, 46, }} 
    \addplot[forget plot, 
    nodes near coords style={yshift=-2pt},
    fill=pink!30!white,
    draw=none, bar shift=.5*\barwidth]
      coordinates {(x0,21) (x1,25) (x3,20) (x5,4) (x7,5) (x10,0)};
      
    \def\CustomLabelsGPT{{23, 76, , , , }} 
    \addplot[forget plot,
    nodes near coords style={yshift=-2pt},
    fill=orange!20!white, draw=none, bar shift=1.5*\barwidth]
      coordinates {(x0,29) (x1,25) (x3,20) (x5,19) (x7,14) (x10,7)};
    \def\CustomLabelsGPTOld{{23, 76, , , , }} 
    \addplot[forget plot,
    nodes near coords style={yshift=-2pt},
    fill=gray!20!white, draw=none, bar shift=2.5*\barwidth]
      coordinates {(x0,22) (x1,5) (x3,0) (x5,0) (x7,0) (x10,0)};
      
      \addplot[fill=red!60!black, bar shift=-2.5*\barwidth, draw=none]
        coordinates {(x0,22) (x1,20) (x3,6) (x5,4) (x7,0) (x10,0) };
      \addplot[fill=blue!60!black, bar shift=-1.5*\barwidth, draw=none]
        coordinates {(x0,29) (x1,23) (x3,9) (x5,8) (x7,3) (x10,0) };
      \addplot[fill=green!60!black, bar shift=-.5*\barwidth, draw=none]
        coordinates {(x0,30) (x1,23) (x3,9) (x5,4) (x7,2) (x10,1) };
      \addplot[fill=pink!60!black, bar shift=.5*\barwidth, draw=none]
        coordinates {(x0,17) (x1,4) (x3,3) (x5,1) (x7,0) (x10,0)};
      \addplot[fill=orange!60!black, bar shift=1.5*\barwidth, draw=none]
        coordinates {(x0,29) (x1,23) (x3,15) (x5,11) (x7,6) (x10,4)};
      \addplot[fill=gray!60!black, bar shift=2.5*\barwidth, draw=none]
        coordinates {(x0,16) (x1,1) (x3,0) (x5,0) (x7,0) (x10,0)}; 

      \addplot[fill=red!60!black, bar shift=-2.5*\barwidth, draw=none]
        coordinates {(x0,-1) (x1,-1) (x3,-1) (x5,-1) (x7,-1) (x10,-1) };
      \addplot[fill=blue!60!black, bar shift=-1.5*\barwidth, draw=none]
        coordinates {(x0,-1) (x1,-1) (x3,-1) (x5,-1) (x7,-1) (x10,-1) };
      \addplot[fill=green!60!black, bar shift=-.5*\barwidth, draw=none]
        coordinates {(x0,-1) (x1,-1) (x3,-1) (x5,-1) (x7,-1) (x10,-1) };
      \addplot[fill=pink!60!black, bar shift=.5*\barwidth, draw=none]
        coordinates {(x0,-1) (x1,-1) (x3,-1) (x5,-1) (x7,-1) (x10,-1)};
      \addplot[fill=orange!60!black, bar shift=1.5*\barwidth, draw=none]
        coordinates {(x0,-1) (x1,-1) (x3,-1) (x5,-1) (x7,-1) (x10,-1)};
      \addplot[fill=gray!60!black, bar shift=2.5*\barwidth, draw=none]
        coordinates {(x0,-1) (x1,-1) (x3,-1) (x5,-1) (x7,-1) (x10,-1)}; 
        
    \end{axis}
    
  \end{tikzpicture}}
    \resizebox{.42\linewidth}{!}{  \begin{tikzpicture}
    \begin{axis}[
        width=8cm,height=4cm,
        ymin = -1,
        ymax = 34,
        legend style={
          at={(1,1)},
          anchor=north east,
        },
        title={\textbf{BabyAI}},
        title style={yshift=-5pt},
        xticklabel style={yshift=3pt, font=\footnotesize},
        ylabel style={yshift=-8pt},
        xlabel = {Constraints No (Avg.~Optimal Cost)},
        xlabel style={align=center, font=\footnotesize\bf, yshift=4pt},   
        ytick = {0, 15, 30},
        yticklabels = {0, 0.5, 1}, 
        yticklabel style={text height=5pt, font=\footnotesize},
        symbolic x coords = {x0, x1, x3, x5, x7, x10},
        xticklabels = {0(3), 1(7), 3(12), 5(15), 7(20), 10(21)},
        xtick={x0, x1, x3, x5, x7, x10}, 
        xtick style={draw=none},
        ybar=0pt,
        bar width=\barwidth, 
        xtick distance=1,               
        enlarge x limits=0.1, 
      ]

    \def\CustomLabelsR{{ , , , , , }} 
    \addplot[forget plot,
      nodes near coords style={yshift=-2pt},
      fill=red!15!white, draw=none, bar shift=-2.5*\barwidth]
      coordinates {(x0,29) (x1,17) (x3,14) (x5,7) (x7,4) (x10,4)};

    \def\CustomLabelsO{{ , , , , , }} 
    \addplot[forget plot,
   nodes near coords style={yshift=-2pt},
   fill=blue!15!white, draw=none, bar shift=-1.5*\barwidth]
      coordinates {(x0,30) (x3,15) (x1,19) (x5,17) (x7,12) (x10,4)};

    \addplot[forget plot,
    nodes near coords style={yshift=-2pt},
    fill=green!15!white, draw=none, bar shift=-.5*\barwidth]
      coordinates {(x0,30) (x1,20) (x3,20) (x5,17) (x7,17) (x10,12)};
    \addplot[forget plot, 
    nodes near coords style={yshift=-2pt},
    fill=pink!30!white,
    draw=none, bar shift=.5*\barwidth]
      coordinates {(x0,16) (x1,5) (x3,11) (x5,4) (x7,10) (x10,9)};
      
    \addplot[forget plot,
    nodes near coords style={yshift=-2pt},
    fill=orange!20!white, draw=none, bar shift=1.5*\barwidth]
      coordinates {(x0,27) (x1,19) (x3,4) (x5,6) (x7,5) (x10,6)};

    \addplot[forget plot,
    nodes near coords style={yshift=-2pt},
    fill=gray!20!white, draw=none, bar shift=2.5*\barwidth]
      coordinates {(x0,27) (x1,11) (x3,0) (x5,0) (x7,0) (x10,0)}; 
      
      \addplot[fill=red!60!black, bar shift=-2.5*\barwidth, draw=none]
        coordinates {(x0,29) (x1,11) (x3,14) (x5,2) (x7,2) (x10,0) };
      \addplot[fill=blue!60!black, bar shift=-1.5*\barwidth, draw=none]
        coordinates {(x0,30) (x1,14) (x3,16) (x5,9) (x7,6) (x10,4) };
      \addplot[fill=green!60!black, bar shift=-.5*\barwidth, draw=none]
        coordinates {(x0,30) (x1,13) (x3,14) (x5,6) (x7,3) (x10,1) };
      \addplot[fill=pink!60!black, bar shift=.5*\barwidth, draw=none]
        coordinates {(x0,16) (x1,4) (x3,2) (x5,1) (x7,2) (x10,0)};
      \addplot[fill=orange!60!black, bar shift=1.5*\barwidth, draw=none]
        coordinates {(x0,30) (x1,19) (x3,4) (x5,5) (x7,5) (x10,5)};
      \addplot[fill=gray!60!black, bar shift=2.5*\barwidth, draw=none]
        coordinates {(x0,27) (x1,7) (x3,0) (x5,0) (x7,0) (x10,0)}; 

      \addplot[fill=red!60!black, bar shift=-2.5*\barwidth, draw=none]
        coordinates {(x0,-1) (x1,-1) (x3,-1) (x5,-1) (x7,-1) (x10,-1) };
      \addplot[fill=blue!60!black, bar shift=-1.5*\barwidth, draw=none]
        coordinates {(x0,-1) (x1,-1) (x3,-1) (x5,-1) (x7,-1) (x10,-1) };
      \addplot[fill=green!60!black, bar shift=-.5*\barwidth, draw=none]
        coordinates {(x0,-1) (x1,-1) (x3,-1) (x5,-1) (x7,-1) (x10,-1) };
      \addplot[fill=pink!60!black, bar shift=.5*\barwidth, draw=none]
        coordinates {(x0,-1) (x1,-1) (x3,-1) (x5,-1) (x7,-1) (x10,-1)};
      \addplot[fill=orange!60!black, bar shift=1.5*\barwidth, draw=none]
        coordinates {(x0,-1) (x1,-1) (x3,-1) (x5,-1) (x7,-1) (x10,-1)};
      \addplot[fill=gray!60!black, bar shift=2.5*\barwidth, draw=none]
        coordinates {(x0,-1) (x1,-1) (x3,-1) (x5,-1) (x7,-1) (x10,-1)}; 
        
    \end{axis}
    
  \end{tikzpicture}}
    \resizebox{.45\linewidth}{!}{  \begin{tikzpicture}
    \begin{axis}[
        width=8cm,height=4cm,
        ymin = -1,
        ymax = 34,
        legend style={
          at={(1,1)},
          anchor=north east,
        },
        title={\textbf{Logistics}},
        title style={yshift=-5pt},
        xticklabel style={yshift=3pt, font=\footnotesize},
        ylabel style={yshift=-8pt},
        ytick = {0, 15, 30},
        yticklabels = {0, 0.5, 1}, 
        yticklabel style={text height=5pt, font=\footnotesize},
        xlabel = {Constraints No (Avg.~Optimal Cost)},
        xlabel style={align=center, font=\footnotesize\bf, yshift=4pt},   
        symbolic x coords = {x0, x1, x3, x5, x7, x10},
        ylabel = {Performance},
        ylabel style={align=center, font=\footnotesize\bf, yshift=-4pt}, 
        xticklabels = {0(7), 1(17), 3(23), 5(27), 7(32), 10(37)},
        xtick={x0, x1, x3, x5, x7, x10}, 
        xtick style={draw=none},
        ybar=0pt,
        bar width=\barwidth, 
        xtick distance=1,               
        enlarge x limits=0.1, 
      ]

    \addplot[forget plot,
      fill=red!15!white, draw=none, bar shift=-2.5*\barwidth]
      coordinates {(x0,24) (x1,23) (x3,12) (x5,1) (x7,1) (x10,1)};

    \addplot[forget plot,
            fill=blue!15!white, draw=none, bar shift=-1.5*\barwidth]
      coordinates {(x0,30) (x1,28) (x3,26) (x5,21) (x7,18) (x10,12)};
      
    \addplot[forget plot,
    fill=green!15!white, draw=none, bar shift=-.5*\barwidth]
      coordinates {(x0,30) (x1,29) (x3,23) (x5,22) (x7,23) (x10,15)};

    \addplot[forget plot, 
        nodes near coords style={yshift=-2pt},
        fill=pink!30!white,
        draw=none, bar shift=.5*\barwidth]
      coordinates {(x0,25) (x1,13) (x3,12) (x5,10) (x7,7) (x10,0)};

    \addplot[forget plot,
    nodes near coords style={yshift=-2pt},
    fill=orange!20!white, draw=none, bar shift=1.5*\barwidth]
      coordinates {(x0,28) (x1,29) (x3,27) (x5,26) (x7,21) (x10,19)};

    \addplot[forget plot,
    nodes near coords style={yshift=-2pt},
    fill=gray!20!white, draw=none, bar shift=2.5*\barwidth]
      coordinates {(x0,5) (x1,1) (x3,0) (x5,0) (x7,0) (x10,0)}; 
      
      \addplot[fill=red!60!black, bar shift=-2.5*\barwidth, draw=none]
        coordinates {(x0,22) (x1,16) (x3,5) (x5,0) (x7,0) (x10,2) };
      \addplot[fill=blue!60!black, bar shift=-1.5*\barwidth, draw=none]
        coordinates {(x0,30) (x1,24) (x3,22) (x5,14) (x7,2) (x10,6) };
      \addplot[fill=green!60!black, bar shift=-.5*\barwidth, draw=none]
        coordinates {(x0,29) (x1,17) (x3,7) (x5,2) (x7,1) (x10,2) };
      \addplot[fill=pink!60!black, bar shift=.5*\barwidth, draw=none]
        coordinates {(x0,23) (x1,7) (x3,3) (x5,1) (x7,0) (x10,0)};
      \addplot[fill=orange!60!black, bar shift=1.5*\barwidth, draw=none]
        coordinates {(x0,28) (x1,28) (x3,25) (x5,17) (x7,5) (x10,6)};
      \addplot[fill=gray!60!black, bar shift=2.5*\barwidth, draw=none]
        coordinates {(x0,3) (x1,0) (x3,0) (x5,0) (x7,0) (x10,0)}; 

      \addplot[fill=red!60!black, bar shift=-2.5*\barwidth, draw=none]
        coordinates {(x0,-1) (x1,-1) (x3,-1) (x5,-1) (x7,-1) (x10,-1) };
      \addplot[fill=blue!60!black, bar shift=-1.5*\barwidth, draw=none]
        coordinates {(x0,-1) (x1,-1) (x3,-1) (x5,-1) (x7,-1) (x10,-1) };
      \addplot[fill=green!60!black, bar shift=-.5*\barwidth, draw=none]
        coordinates {(x0,-1) (x1,-1) (x3,-1) (x5,-1) (x7,-1) (x10,-1) };
      \addplot[fill=pink!60!black, bar shift=.5*\barwidth, draw=none]
        coordinates {(x0,-1) (x1,-1) (x3,-1) (x5,-1) (x7,-1) (x10,-1)};
      \addplot[fill=orange!60!black, bar shift=1.5*\barwidth, draw=none]
        coordinates {(x0,-1) (x1,-1) (x3,-1) (x5,-1) (x7,-1) (x10,-1)};
      \addplot[fill=gray!60!black, bar shift=2.5*\barwidth, draw=none]
        coordinates {(x0,-1) (x1,-1) (x3,-1) (x5,-1) (x7,-1) (x10,-1)}; 
        
    \end{axis}
    
  \end{tikzpicture}}
    \resizebox{.42\linewidth}{!}{  \begin{tikzpicture}
    \begin{axis}[
        width=8cm,height=4cm,
        ymin = -1,
        ymax = 34,
        legend style={
          at={(1,1)},
          anchor=north east,
        },
        title={\textbf{Sokoban}},
        title style={yshift=-5pt},
        xticklabel style={yshift=3pt, font=\footnotesize},
        ylabel style={yshift=-8pt},
        ytick = {0, 15, 30},
        yticklabels = {0, 0.5, 1}, 
        yticklabel style={text height=5pt, font=\footnotesize},
        xlabel = {Constraints No (Avg.~Optimal Cost)},
        xlabel style={align=center, font=\footnotesize\bf, yshift=4pt},   
        symbolic x coords = {x0, x1, x3, x5, x7, x10},
        xticklabels = {0(7), 1(11), 3(13), 5(16), 7(19), 10(24)}, 
        xtick={x0, x1, x3, x5, x7, x10}, 
        xtick style={draw=none},
        ybar=0pt,
        bar width=\barwidth, 
        xtick distance=1,               
        enlarge x limits=0.1, 
      ]

    \def\CustomLabelsR{{ , , , , , }} 
    \addplot[forget plot,
     nodes near coords={%
        \pgfmathsetmacro{\mylabel}{\CustomLabelsR[\coordindex]}
        \notsotiny\mylabel
      },
      nodes near coords style={yshift=-2pt},
      fill=red!15!white, draw=none, bar shift=-2.5*\barwidth]
      coordinates {(x0,18) (x1,12) (x3,6) (x5,3) (x7,0) (x10,0)};

    \def\CustomLabelsO{{ , , , , , }} 
    \addplot[forget plot,
     nodes near coords={%
        \pgfmathsetmacro{\mylabel}{\CustomLabelsO[\coordindex]}
        \hspace{-2pt}\notsotiny\mylabel
   },
   nodes near coords style={yshift=-2pt},
   fill=blue!15!white, draw=none, bar shift=-1.5*\barwidth]
         coordinates {(x0,29) (x1,25) (x3,24) (x5,21) (x7,9) (x10,7)};
      
    \def\CustomLabelsG{{ , , , , , }} 
    \addplot[forget plot,
     nodes near coords={%
        \pgfmathsetmacro{\mylabel}{\CustomLabelsG[\coordindex]}
        \hspace{2pt}\notsotiny\mylabel
    },
    nodes near coords style={yshift=-2pt},
    fill=green!15!white, draw=none, bar shift=-.5*\barwidth]
      coordinates {(x0,29) (x1,23) (x3,24) (x5,12) (x7,12) (x10,1)};

    \def\CustomLabelsC{{ , , , , , }} 
    \addplot[forget plot, 
     nodes near coords={%
        \pgfmathsetmacro{\mylabel}{\CustomLabelsC[\coordindex]}
        \hspace{2pt}\notsotiny\mylabel
    },
    nodes near coords style={yshift=-2pt},
    fill=pink!30!white,
    draw=none, bar shift=.5*\barwidth]
      coordinates {(x0,0) (x1,4) (x3,4) (x5,2) (x7,0) (x10,8)};
      
    \def\CustomLabelsGPT{{ , , , , , }} 
    \addplot[forget plot,
     nodes near coords={%
        \pgfmathsetmacro{\mylabel}{\CustomLabelsGPT[\coordindex]}
        \hspace{2pt}\notsotiny\mylabel
    },
    nodes near coords style={yshift=-2pt},
    fill=orange!20!white, draw=none, bar shift=1.5*\barwidth]
      coordinates {(x0,30) (x1,27) (x3,25) (x5,20) (x7,19) (x10,8)};

    \def\CustomLabelsGPTOld{{ , , , , , }} 
    \addplot[forget plot,
     nodes near coords={%
        \pgfmathsetmacro{\mylabel}{\CustomLabelsGPT[\coordindex]}
        \hspace{2pt}\notsotiny\mylabel
    },
    nodes near coords style={yshift=-2pt},
    fill=gray!20!white, draw=none, bar shift=2.5*\barwidth]
      coordinates {(x0,0) (x1,0) (x3,0) (x5,0) (x7,0) (x10,0)};
      
      \addplot[fill=red!60!black, bar shift=-2.5*\barwidth, draw=none]
        coordinates {(x0,13) (x1,12) (x3,1) (x5,0) (x7,0) (x10,0) };
      \addplot[fill=blue!60!black, bar shift=-1.5*\barwidth, draw=none]
        coordinates {(x0,28) (x1,20) (x3,15) (x5,12) (x7,4) (x10,0) };
      \addplot[fill=green!60!black, bar shift=-.5*\barwidth, draw=none]
        coordinates {(x0,28) (x1,20) (x3,6) (x5,3) (x7,6) (x10,0) };
      \addplot[fill=pink!60!black, bar shift=.5*\barwidth, draw=none]
        coordinates {(x0,0) (x1,4) (x3,0) (x5,0) (x7,0) (x10,0)};
      \addplot[fill=orange!60!black, bar shift=1.5*\barwidth, draw=none]
        coordinates {(x0,30) (x1,24) (x3,19) (x5,16) (x7,8) (x10,6)};
      \addplot[fill=gray!60!black, bar shift=2.5*\barwidth, draw=none]
        coordinates {(x0,0) (x1,0) (x3,0) (x5,0) (x7,0) (x10,0)}; 

      \addplot[fill=red!60!black, bar shift=-2.5*\barwidth, draw=none]
        coordinates {(x0,-1) (x1,-1) (x3,-1) (x5,-1) (x7,-1) (x10,-1) };
      \addplot[fill=blue!60!black, bar shift=-1.5*\barwidth, draw=none]
        coordinates {(x0,-1) (x1,-1) (x3,-1) (x5,-1) (x7,-1) (x10,-1) };
      \addplot[fill=green!60!black, bar shift=-.5*\barwidth, draw=none]
        coordinates {(x0,-1) (x1,-1) (x3,-1) (x5,-1) (x7,-1) (x10,-1) };
      \addplot[fill=pink!60!black, bar shift=.5*\barwidth, draw=none]
        coordinates {(x0,-1) (x1,-1) (x3,-1) (x5,-1) (x7,-1) (x10,-1)};
      \addplot[fill=orange!60!black, bar shift=1.5*\barwidth, draw=none]
        coordinates {(x0,-1) (x1,-1) (x3,-1) (x5,-1) (x7,-1) (x10,-1)};
      \addplot[fill=gray!60!black, bar shift=2.5*\barwidth, draw=none]
        coordinates {(x0,-1) (x1,-1) (x3,-1) (x5,-1) (x7,-1) (x10,-1)};

    \end{axis}
    
  \end{tikzpicture}}
    \resizebox{.45\linewidth}{!}{  \begin{tikzpicture}
    \begin{axis}[
        width=8cm,height=4cm,
        ymin = -1,
        ymax = 34,
        legend style={
          at={(1,1)},
          anchor=north east,
        },
        title={\textbf{AlfWorld}},
        title style={yshift=-5pt},
        xticklabel style={yshift=3pt, font=\footnotesize},
        ylabel style={yshift=-8pt},
        ytick = {0, 15, 30},
        yticklabels = {0, 0.5, 1}, 
        yticklabel style={text height=5pt, font=\footnotesize},
        ylabel = {Performance},
        ylabel style={align=center, font=\footnotesize\bf, yshift=-4pt},  
        xlabel = {Constraints No (Avg.~Optimal Cost)},
        xlabel style={align=center, font=\footnotesize\bf, yshift=4pt},  
        symbolic x coords = {x0, x1, x3, x5, x7, x10},
        xticklabels = {0(4), 1(7), 3(11), 5(15), 7(18), 10(22)}, 
        xtick={x0, x1, x3, x5, x7, x10}, 
        xtick style={draw=none},
        ybar=0pt,
        bar width=\barwidth, 
        xtick distance=1,               
        enlarge x limits=0.1, 
      ]

    \def\CustomLabelsR{{ , , , , , }} 
    \addplot[forget plot,
     nodes near coords={%
        \pgfmathsetmacro{\mylabel}{\CustomLabelsR[\coordindex]}
        \notsotiny\mylabel
      },
      nodes near coords style={yshift=-2pt},
      fill=red!15!white, draw=none, bar shift=-2.5*\barwidth]
      coordinates {(x0,15) (x1,5) (x3,0) (x5,0) (x7,0) (x10,0)};

    \def\CustomLabelsO{{ , , , , , }} 
    \addplot[forget plot,
     nodes near coords={%
        \pgfmathsetmacro{\mylabel}{\CustomLabelsO[\coordindex]}
        \hspace{-2pt}\notsotiny\mylabel
   },
   nodes near coords style={yshift=-2pt},
   fill=blue!15!white, draw=none, bar shift=-1.5*\barwidth]
         coordinates {(x0,11) (x1,10) (x3,0) (x5,0) (x7,1) (x10,1)};
      
    \def\CustomLabelsG{{ , , , , , }} 
    \addplot[forget plot,
     nodes near coords={%
        \pgfmathsetmacro{\mylabel}{\CustomLabelsG[\coordindex]}
        \hspace{2pt}\notsotiny\mylabel
    },
    nodes near coords style={yshift=-2pt},
    fill=green!15!white, draw=none, bar shift=-.5*\barwidth]
      coordinates {(x0,5) (x1,1) (x3,0) (x5,0) (x7,0) (x10,0)};

    \def\CustomLabelsC{{ , , , , , }} 
    \addplot[forget plot, 
     nodes near coords={%
        \pgfmathsetmacro{\mylabel}{\CustomLabelsC[\coordindex]}
        \hspace{2pt}\notsotiny\mylabel
    },
    nodes near coords style={yshift=-2pt},
    fill=pink!30!white,
    draw=none, bar shift=.5*\barwidth]
      coordinates {(x0,10) (x1,0) (x3,0) (x5,0) (x7,3) (x10,1)};
      
    \def\CustomLabelsGPT{{ , , , , , }} 
    \addplot[forget plot,
     nodes near coords={%
        \pgfmathsetmacro{\mylabel}{\CustomLabelsGPT[\coordindex]}
        \hspace{2pt}\notsotiny\mylabel
    },
    nodes near coords style={yshift=-2pt},
    fill=orange!20!white, draw=none, bar shift=1.5*\barwidth]
      coordinates {(x0,25) (x1,24) (x3,23) (x5,20) (x7,14) (x10,15)};
      
    \def\CustomLabelsGPTOld{{ , , , , , }} 
    \addplot[forget plot,
     nodes near coords={%
        \pgfmathsetmacro{\mylabel}{\CustomLabelsGPT[\coordindex]}
        \hspace{2pt}\notsotiny\mylabel
    },
    nodes near coords style={yshift=-2pt},
    fill=gray!20!white, draw=none, bar shift=2.5*\barwidth]
      coordinates {(x0,10) (x1,0) (x3,0) (x5,0) (x7,0) (x10,0)};
      
      \addplot[fill=red!60!black, bar shift=-2.5*\barwidth, draw=none]
        coordinates {(x0,-1) (x1,-1) (x3,-1) (x5,-1) (x7,-1) (x10,-1) };
      \addplot[fill=blue!60!black, bar shift=-1.5*\barwidth, draw=none]
        coordinates {(x0,-1) (x1,-1) (x3,-1) (x5,-1) (x7,-1) (x10,-1) };
      \addplot[fill=green!60!black, bar shift=-.5*\barwidth, draw=none]
        coordinates {(x0,-1) (x1,-1) (x3,-1) (x5,-1) (x7,-1) (x10,-1) };
      \addplot[fill=pink!60!black, bar shift=.5*\barwidth, draw=none]
        coordinates {(x0,-1) (x1,-1) (x3,-1) (x5,-1) (x7,-1) (x10,-1)};
      \addplot[fill=orange!60!black, bar shift=1.5*\barwidth, draw=none]
        coordinates {(x0,-1) (x1,-1) (x3,-1) (x5,-1) (x7,-1) (x10,-1)};
      \addplot[fill=gray!60!black, bar shift=2.5*\barwidth, draw=none]
        coordinates {(x0,-1) (x1,-1) (x3,-1) (x5,-1) (x7,-1) (x10,-1)};
        
      \addplot[fill=red!60!black, bar shift=-2.5*\barwidth, draw=none]
        coordinates {(x0,15) (x1,5) (x3,0) (x5,0) (x7,0) (x10,0) };
      \addplot[fill=blue!60!black, bar shift=-1.5*\barwidth, draw=none]
        coordinates {(x0,11) (x1,9) (x3,0) (x5,0) (x7,0) (x10,0) };
      \addplot[fill=green!60!black, bar shift=-.5*\barwidth, draw=none]
        coordinates {(x0,5) (x1,1) (x3,0) (x5,0) (x7,0) (x10,0) };
      \addplot[fill=pink!60!black, bar shift=.5*\barwidth, draw=none]
        coordinates {(x0,10) (x1,0) (x3,0) (x5,0) (x7,0) (x10,0)};
      \addplot[fill=orange!60!black, bar shift=1.5*\barwidth, draw=none]
        coordinates {(x0,25) (x1,24) (x3,22) (x5,14) (x7,11) (x10,6)};
      \addplot[fill=gray!60!black, bar shift=2.5*\barwidth, draw=none]
        coordinates {(x0,6) (x1,0) (x3,0) (x5,0) (x7,0) (x10,0)};
    \end{axis}
    
  \end{tikzpicture}}
    \resizebox{\linewidth}{!}{\begin{tikzpicture}
      \begin{customlegend}[legend entries={DeepSeek R1, OpenAI o3, Gemini 2.5 Pro, Claude 3.7 Sonnet, GPT-5, GPT-4.1 CoT}, legend columns=-1, legend style={draw=none}]
          \addlegendimage{red!60!black, area legend, fill}
          \addlegendimage{blue!60!black, area legend, fill}
          \addlegendimage{green!60!black, area legend, fill}
          \addlegendimage{pink!60!black, area legend, fill}
          \addlegendimage{orange!60!black, area legend, fill}
          \addlegendimage{gray!60!black, area legend, fill}
      \end{customlegend}
\end{tikzpicture}}
\caption{\textbf{Performance vs. number of constraints (average optimal cost)}. Performance denotes the percentage of problems solved with an optimal plan (colored bars) or with a valid, but possibly suboptimal plan (background faded-out bars).
}
%
\label{fig:llm_results}
\end{figure}
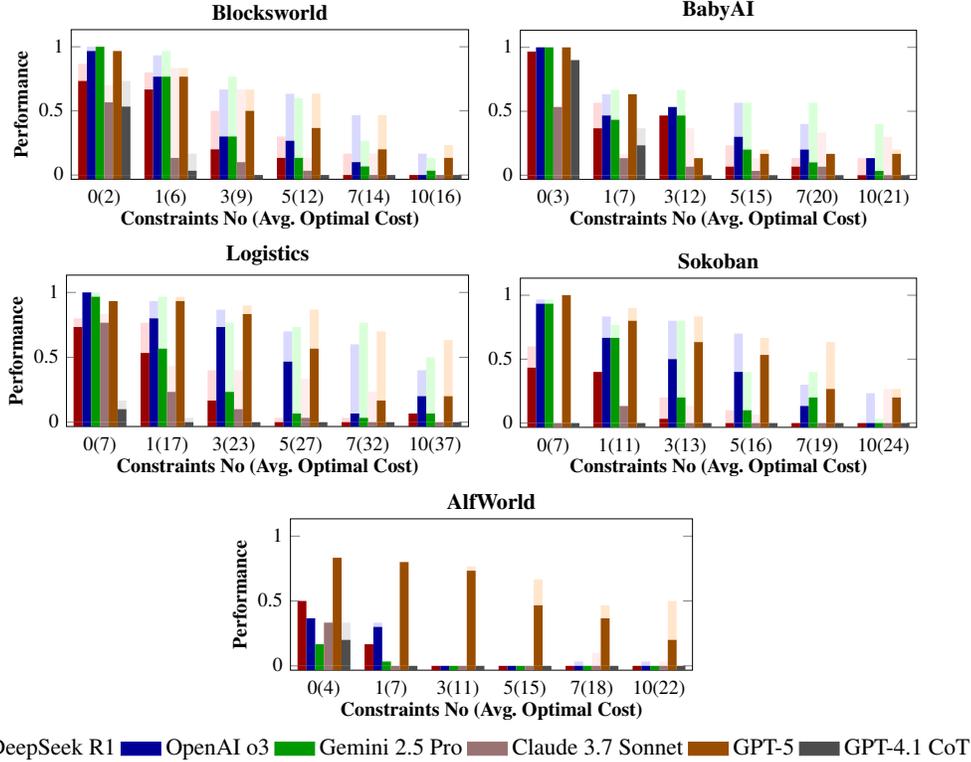
\begin{figure}[t]
    \centering
    \resizebox{.33\linewidth}{!}{\begin{tikzpicture}
\begin{axis}[
    title={\textbf{Blocksworld}},
    title style={yshift=-5pt},
    width=6cm,
    height=3cm,
    only marks,
    mark size=1pt,
    ytick = {0, 10000, 20000, 30000},
    yticklabels = {0, 1, 2, 3}, 
    yticklabel style={text height=5pt, font=\footnotesize},
    ylabel = {Reasoning\\ Tokens},
    ylabel style={align=center, font=\footnotesize\bf, yshift=-13pt}, 
    xtick = {10, 20, 30}, 
    xticklabels = {10, 20, 30}, 
    xticklabel style={text height=5pt, font=\footnotesize},
    xlabel = {Optimal Cost},
    xlabel style={align=center, font=\footnotesize\bf, yshift=8pt}, 
    ymax = 39000,
    scatter/classes={
        deepseek={mark=*,red!60!black},
        o3={mark=*,blue!60!black},
        gemini-2.5={mark=*,green!60!black},
        claude_37_sonnet={mark=*,pink!60!black},
        gpt-5={mark=*,orange!60!black}  
    }
]
\addplot[
    scatter,
    only marks,
    scatter src=explicit symbolic
]
table[x=x, y=y, meta=label] {data/tokens_blocksworld.csv};
\end{axis}
\end{tikzpicture}}
    \resizebox{.3\linewidth}{!}{\begin{tikzpicture}
\begin{axis}[
    title={\textbf{BabyAI}},
    title style={yshift=-5pt},
    width=6cm,
    height=3cm,
    only marks,
    mark size=1pt,
    ytick = {0, 10000, 20000, 30000},
    yticklabels = {0, 1, 2, 3}, 
    yticklabel style={text height=5pt, font=\footnotesize},
    xtick = {10, 20, 30, 40}, 
    xticklabels = {10, 20, 30, 40}, 
    xticklabel style={text height=5pt, font=\footnotesize},
    xlabel = {Optimal Cost},
    xlabel style={align=center, font=\footnotesize\bf, yshift=8pt},   
    ymax = 39000,
    scatter/classes={
        deepseek={mark=*,red!60!black},
        o3={mark=*,blue!60!black},
        gemini-2.5={mark=*,green!60!black},
        claude_37_sonnet={mark=*,pink!60!black},
        gpt-5={mark=*,orange!60!black} 
    }
]
\addplot[
    scatter,
    only marks,
    scatter src=explicit symbolic
]
table[x=x, y=y, meta=label] {data/tokens_babyai.csv};
\end{axis}
\end{tikzpicture}}
    \resizebox{.3\linewidth}{!}{\begin{tikzpicture}
\begin{axis}[
    title={\textbf{Logistics}},
    title style={yshift=-5pt},
    width=6cm,
    height=3cm,
    only marks,
    mark size=1pt,
    ytick = {0, 10000, 20000, 30000},
    yticklabels = {0, 1, 2, 3}, 
    yticklabel style={text height=5pt, font=\footnotesize},
    xtick = {20, 40, 60}, 
    xticklabels = {20 , 40, 60}, 
    xticklabel style={text height=5pt, font=\footnotesize},
    xlabel = {Optimal Cost},
    xlabel style={align=center, font=\footnotesize\bf, yshift=8pt},
    ymax = 39000,
    scatter/classes={
        deepseek={mark=*,red!60!black},
        o3={mark=*,blue!60!black},
        gemini-2.5={mark=*,green!60!black},
        claude_37_sonnet={mark=*,pink!60!black},
        gpt-5={mark=*,orange!60!black} 
    }
]
\addplot[
    scatter,
    only marks,
    scatter src=explicit symbolic
]
table[x=x, y=y, meta=label] {data/tokens_logistics.csv};
\end{axis}
\end{tikzpicture}}
    \hspace{5pt}
    \resizebox{.33\linewidth}{!}{\begin{tikzpicture}
\begin{axis}[
    title={\textbf{Sokoban}},
    title style={yshift=-5pt},
    width=6cm,
    height=3cm,
    only marks,
    mark size=1pt,
    ytick = {0, 10000, 20000, 30000},
    yticklabels = {0, 1, 2, 3}, 
    yticklabel style={text height=5pt, font=\footnotesize},
    ylabel = {Reasoning\\ Tokens},
    ylabel style={align=center, font=\footnotesize\bf, yshift=-13pt},  
    xtick = {10, 20, 30}, 
    xticklabels = {10, 20, 30}, 
    xticklabel style={text height=5pt, font=\footnotesize},
    xlabel = {Optimal Cost},
    xlabel style={align=center, font=\footnotesize\bf, yshift=8pt},  
    ymax = 39000,
    scatter/classes={
        deepseek={mark=*,red!60!black},
        o3={mark=*,blue!60!black},
        gemini-2.5={mark=*,green!60!black},
        claude_37_sonnet={mark=*,pink!60!black},
        gpt-5={mark=*,orange!60!black} 
    }
]
\addplot[
   scatter,
    only marks,
    scatter src=explicit symbolic
]
table[x=x, y=y, meta=label] {data/tokens_sokoban.csv};
\end{axis}
\end{tikzpicture}}
    \resizebox{.3\linewidth}{!}{\begin{tikzpicture}
\begin{axis}[
    title={\textbf{AlfWorld}},
    title style={yshift=-5pt},
    width=6cm,
    height=3cm,
    only marks,
    mark size=1pt,
    ytick = {0, 10000, 20000, 30000},
    yticklabels = {0, 1, 2, 3}, 
    yticklabel style={text height=5pt, font=\footnotesize},
    xtick = {10, 20, 30}, 
    xticklabels = {10, 20, 30}, 
    xticklabel style={text height=5pt, font=\footnotesize},
    xlabel = {Optimal Cost},
    xlabel style={align=center, font=\footnotesize\bf, yshift=8pt}, 
    ymax = 39000,
    scatter/classes={
        deepseek={mark=*,red!60!black},
        o3={mark=*,blue!60!black},
        gemini-2.5={mark=*,green!60!black},
        claude_37_sonnet={mark=*,pink!60!black},
        gpt-5={mark=*,orange!60!black}
    }
]
\addplot[
    scatter,
    only marks,
    scatter src=explicit symbolic
]
table[x=x, y=y, meta=label] {data/tokens_alfworld.csv};
\end{axis}
\end{tikzpicture}}
    \begin{tikzpicture}
      \begin{customlegend}[legend entries={DeepSeek R1, OpenAI o3, Gemini 2.5 Pro, Claude 3.7 Sonnet, GPT-5}, legend columns=-1, legend style={draw=none}]
          \addlegendimage{red!60!black, area legend, fill}
          \addlegendimage{blue!60!black, area legend, fill}
          \addlegendimage{green!60!black, area legend, fill}
          \addlegendimage{pink!60!black, area legend, fill}
          \addlegendimage{orange!60!black, area legend, fill} 
      \end{customlegend}
\end{tikzpicture}
\caption{\textbf{Correlation between Reasoning Tokens and Optimal Cost}.
The Pearson correlation coefficients are as follows. 
\textbf{Blocksworld:} [\textcolor{pink!60!black}{0.48}, \textcolor{red!60!black}{0.6}, \textcolor{green!60!black}{0.67}, \textcolor{blue!60!black}{0.66}, \textcolor{orange!60!black}{0.63}]; \textbf{BabyAI:} [\textcolor{pink!60!black}{0.52}, \textcolor{red!60!black}{0.68}, \textcolor{green!60!black}{0.7}, \textcolor{blue!60!black}{0.74}, \textcolor{orange!60!black}{0.8}]; \textbf{Logistics:} [\textcolor{pink!60!black}{0.16}, \textcolor{red!60!black}{0.37}, \textcolor{green!60!black}{0.73}, \textcolor{blue!60!black}{0.71}, \textcolor{orange!60!black}{0.77}]; \textbf{Sokoban:} [\textcolor{pink!60!black}{0.53}, \textcolor{red!60!black}{0.12}, \textcolor{green!60!black}{0.57}, \textcolor{blue!60!black}{0.62}, \textcolor{orange!60!black}{0.72}]; \textbf{AlfWorld:} [\textcolor{pink!60!black}{0.46}, \textcolor{red!60!black}{0.74}, \textcolor{green!60!black}{0.86}, \textcolor{blue!60!black}{0.55}, \textcolor{orange!60!black}{0.87}].
} 
\label{fig:benchmark_stats}
\end{figure}
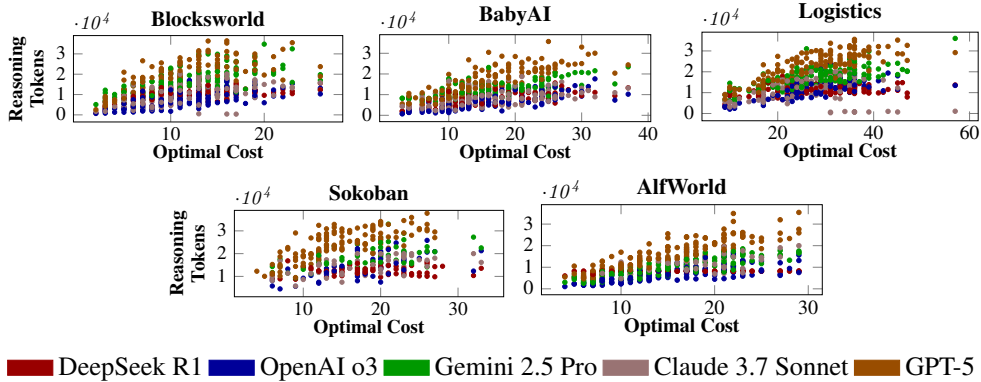
\begin{figure*}[t]
    \centering
    \includegraphics[width=0.9\linewidth]{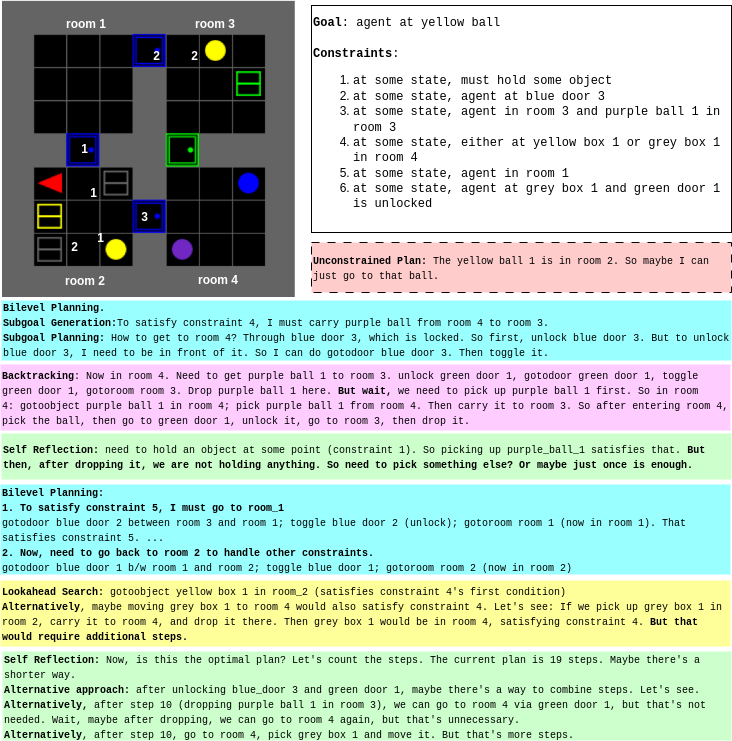}
    \captionof{figure}{\textbf{Planning Traces from R1}. The white box displays the goal and constraints. On the left is the initial observation from the BabyAI environment. Colored boxes indicate model behaviours: \textcolor{cyan}{cyan} for bilevel planning, \textcolor{Chartreuse3}{green} for self-reflection, \textcolor{Goldenrod2}{yellow} for lookahead search, and \textcolor{violet}{violet} backtracking. We also show the 1-step unconstrained plan generated by R1 for the same goal in \textcolor{Salmon}{salmon - - -}.}
    \label{fig:r1-annotated-trace}
\end{figure*}
\begin{figure}[t]
    \centering
    \resizebox{.45\linewidth}{!}{
      \begin{tikzpicture}
    \begin{axis}[
        width=8cm,height=4cm,
        ymin = 1,
        ymode = log,
        legend style={
          at={(1,1)},
          anchor=north east,
        },
        title={\textbf{Blocksworld}},
        title style={yshift=-5pt},
        xticklabel style={yshift=3pt, font=\footnotesize},
        ylabel style={yshift=-8pt},
        ytick = {1, 10, 100, 1000},
        yticklabels = {1, 10, 100, }, 
        xlabel = {Constraints No (Avg.~Optimal Cost)},
        xlabel style={align=center, font=\footnotesize\bf, yshift=4pt},   
        yticklabel style={text height=5pt, font=\footnotesize},
        ylabel = {Execution Time},
        ylabel style={align=center, font=\footnotesize\bf, yshift=-5pt}, 
        ymax = 1100,
        symbolic x coords = {x1, x3, x5, x7, x10},
        xticklabels = {1(6), 3(9), 5(12), 7(14), 10(16)},
        xtick={x1, x3, x5, x7, x10}, 
        xtick style={draw=none},
        ybar=0pt,
        bar width=\barwidth, 
        xtick distance=1,               
        enlarge x limits=0.1, 
      ]

      \addplot[fill=yellow!60!black, bar shift=-2.5*\barwidth, draw=none]
        coordinates {(x1,1.4) (x3,4) (x5,9) (x7,20) (x10,45) };
      \addplot[fill=red!60!black, bar shift=-1.5*\barwidth, draw=none]
        coordinates {(x1,269) (x3,440) (x5,394) (x7,509) (x10,479) };
      \addplot[fill=blue!60!black, bar shift=-.5*\barwidth, draw=none]
        coordinates {(x1,44) (x3,200) (x5,294) (x7,157) (x10,352) };
      \addplot[fill=green!60!black, bar shift=.5*\barwidth, draw=none]
        coordinates {(x1,90) (x3,129) (x5,157) (x7,170) (x10,151) };
      \addplot[fill=pink!60!black, bar shift=1.5*\barwidth, draw=none]
        coordinates {(x1,76) (x3,121) (x5,120) (x7,151) (x10,167)};
      \addplot[fill=orange!60!black, bar shift=2.5*\barwidth, draw=none]
        coordinates {(x1,157) (x3,286) (x5,697) (x7,503) (x10,510)};
        
    \end{axis}
    
  \end{tikzpicture}}
    \resizebox{.42\linewidth}{!}{  \begin{tikzpicture}
    \begin{axis}[
        width=8cm,height=4cm,
        ymin = 1,
        legend style={
          at={(1,1)},
          anchor=north east,
        },
        ymode = log,
        title={\textbf{BabyAI}},
        title style={yshift=-5pt},
        xticklabel style={yshift=3pt, font=\footnotesize},
        ylabel style={yshift=-8pt},
        xlabel = {Constraints No (Avg.~Optimal Cost)},
        xlabel style={align=center, font=\footnotesize\bf, yshift=4pt},   
        ytick = {1, 10, 100, 1000},
        yticklabels = {1, 10, 100, }, 
        yticklabel style={text height=5pt, font=\footnotesize},
        ymax = 1100, 
        symbolic x coords = {x1, x3, x5, x7, x10},
        xticklabels = {1(7), 3(12), 5(15), 7(20), 10(21)},
        xtick={x1, x3, x5, x7, x10}, 
        xtick style={draw=none},
        ybar=0pt,
        bar width=\barwidth, 
        xtick distance=1,               
        enlarge x limits=0.1, 
      ]

      \addplot[fill=yellow!60!black, bar shift=-2.5*\barwidth, draw=none]
        coordinates {(x1,2) (x3,4) (x5,9) (x7,8) (x10,31) };
      \addplot[fill=red!60!black, bar shift=-1.5*\barwidth, draw=none]
        coordinates {(x1,185) (x3,306) (x5,386) (x7,434) (x10,446) };
      \addplot[fill=blue!60!black, bar shift=-.5*\barwidth, draw=none]
        coordinates {(x1,60) (x3,115) (x5,135) (x7,169) (x10,150) };
      \addplot[fill=green!60!black, bar shift=.5*\barwidth, draw=none]
        coordinates {(x1,80) (x3,100) (x5,123) (x7,130) (x10,141) };
      \addplot[fill=pink!60!black, bar shift=1.5*\barwidth, draw=none]
        coordinates {(x1,72) (x3,82) (x5,92) (x7,116) (x10,114)};
      \addplot[fill=orange!60!black, bar shift=2.5*\barwidth, draw=none]
        coordinates {(x1,130) (x3,340) (x5,589) (x7,501) (x10,454)};
        
    \end{axis}
    
  \end{tikzpicture}}
    \resizebox{.45\linewidth}{!}{  \begin{tikzpicture}
    \begin{axis}[
        width=8cm,height=4cm,
        ymin = 1,
        ymode = log,
        legend style={
          at={(1,1)},
          anchor=north east,
        },
        title={\textbf{Logistics}},
        title style={yshift=-5pt},
        xticklabel style={yshift=3pt, font=\footnotesize},
        ylabel style={yshift=-8pt},
        ytick = {1, 10, 100, 1000},
        yticklabels = {1, 10, 100, }, 
        yticklabel style={text height=5pt, font=\footnotesize},
        ylabel = {Execution Time},
        ylabel style={align=center, font=\footnotesize\bf, yshift=-5pt}, 
        xlabel = {Constraints No (Avg.~Optimal Cost)},
        xlabel style={align=center, font=\footnotesize\bf, yshift=4pt},   
        ymax = 1100,
        symbolic x coords = {x1, x3, x5, x7, x10},
        xticklabels = {1(17), 3(23), 5(27), 7(32), 10(37)},
        xtick={x1, x3, x5, x7, x10}, 
        xtick style={draw=none},
        ybar=0pt,
        bar width=\barwidth, 
        xtick distance=1,               
        enlarge x limits=0.1, 
      ]

      \addplot[fill=yellow!60!black, bar shift=-2.5*\barwidth, draw=none]
        coordinates {(x1,1.6) (x3,9) (x5,15) (x7,23) (x10,30) };
      \addplot[fill=red!60!black, bar shift=-1.5*\barwidth, draw=none]
        coordinates {(x1,350) (x3,375) (x5,491) (x7,466) (x10,459) };
      \addplot[fill=blue!60!black, bar shift=-.5*\barwidth, draw=none]
        coordinates {(x1,114) (x3,284) (x5,220) (x7,350) (x10,481) };
      \addplot[fill=green!60!black, bar shift=.5*\barwidth, draw=none]
        coordinates {(x1,136) (x3,129) (x5,153) (x7,196) (x10,250) };
      \addplot[fill=pink!60!black, bar shift=1.5*\barwidth, draw=none]
        coordinates {(x1,100) (x3,127) (x5,182) (x7,179) (x10,230)};
      \addplot[fill=orange!60!black, bar shift=2.5*\barwidth, draw=none]
        coordinates {(x1,310) (x3,366) (x5,455) (x7,479) (x10,637)};
        
    \end{axis}
    
  \end{tikzpicture}}
    \resizebox{.42\linewidth}{!}{  \begin{tikzpicture}
    \begin{axis}[
        width=8cm,height=4cm,
        ymin = 1,
        ymode = log,
        legend style={
          at={(1,1)},
          anchor=north east,
        },
        title={\textbf{Sokoban}},
        title style={yshift=-5pt},
        xticklabel style={yshift=3pt, font=\footnotesize},
        ylabel style={yshift=-8pt},
        ytick = {1, 10, 100, 1000},
        yticklabels = {1, 10, 100, }, 
        yticklabel style={text height=5pt, font=\footnotesize},
        ylabel style={align=center, font=\footnotesize\bf, yshift=-5pt}, 
        ymax = 1100,
        symbolic x coords = {x1, x3, x5, x7, x10},
        xlabel = {Constraints No (Avg.~Optimal Cost)},
        xlabel style={align=center, font=\footnotesize\bf, yshift=4pt},   
        xticklabels = {1(11), 3(13), 5(16), 7(19), 10(24)},
        xtick={x1, x3, x5, x7, x10}, 
        xtick style={draw=none},
        ybar=0pt,
        bar width=\barwidth, 
        xtick distance=1,               
        enlarge x limits=0.1, 
      ]

      \addplot[fill=yellow!60!black, bar shift=-2.5*\barwidth, draw=none]
        coordinates {(x1,4) (x3,6) (x5,13) (x7,33) (x10,62) };
      \addplot[fill=red!60!black, bar shift=-1.5*\barwidth, draw=none]
        coordinates {(x1,371) (x3,415) (x5,374) (x7,400) (x10,430) };
      \addplot[fill=blue!60!black, bar shift=-.5*\barwidth, draw=none]
        coordinates {(x1,200) (x3,220) (x5,236) (x7,383) (x10,665) };
      \addplot[fill=green!60!black, bar shift=.5*\barwidth, draw=none]
        coordinates {(x1,117) (x3,151) (x5,152) (x7,214) (x10,250) };
      \addplot[fill=pink!60!black, bar shift=1.5*\barwidth, draw=none]
        coordinates {(x1,149) (x3,174) (x5,187) (x7,200) (x10,213)};
      \addplot[fill=orange!60!black, bar shift=2.5*\barwidth, draw=none]
        coordinates {(x1,316) (x3,654) (x5,683) (x7,1061) (x10,917)};
        
    \end{axis}
    
  \end{tikzpicture}}
    \resizebox{.45\linewidth}{!}{  \begin{tikzpicture}
    \begin{axis}[
        width=8cm,height=4cm,
        ymin = 1,
        ymode = log,
        legend style={
          at={(1,1)},
          anchor=north east,
        },
        title={\textbf{AlfWorld}},
        title style={yshift=-5pt},
        xticklabel style={yshift=3pt, font=\footnotesize},
        ylabel style={yshift=-8pt},
        xlabel = {Constraints No (Avg.~Optimal Cost)},
        xlabel style={align=center, font=\footnotesize\bf, yshift=4pt},   
        ytick = {1, 10, 100, 1000},
        yticklabels = {1, 10, 100, }, 
        yticklabel style={text height=5pt, font=\footnotesize},
        ylabel = {Execution Time},
        ylabel style={align=center, font=\footnotesize\bf, yshift=-5pt}, 
        ymax = 1100,
        symbolic x coords = {x1, x3, x5, x7, x10},
        xticklabels = {1(7), 3(11), 5(15), 7(18), 10(22)},
        xtick={x1, x3, x5, x7, x10}, 
        xtick style={draw=none},
        ybar=0pt,
        bar width=\barwidth, 
        xtick distance=1,               
        enlarge x limits=0.1, 
      ]

      \addplot[fill=yellow!60!black, bar shift=-2.5*\barwidth, draw=none]
        coordinates {(x1,2.6) (x3,2.5) (x5,3.2) (x7,3) (x10,6) };
      \addplot[fill=red!60!black, bar shift=-1.5*\barwidth, draw=none]
        coordinates {(x1,364) (x3,408) (x5,450) (x7,416) (x10,447) };
      \addplot[fill=blue!60!black, bar shift=-.5*\barwidth, draw=none]
        coordinates {(x1,97) (x3,188) (x5,222) (x7,159) (x10,239) };
      \addplot[fill=green!60!black, bar shift=.5*\barwidth, draw=none]
        coordinates {(x1,59) (x3,94) (x5,109) (x7,113) (x10,132) };
      \addplot[fill=pink!60!black, bar shift=1.5*\barwidth, draw=none]
        coordinates {(x1,25) (x3,129) (x5,151) (x7,135) (x10,177)
        };
      \addplot[fill=orange!60!black, bar shift=2.5*\barwidth, draw=none]
        coordinates {(x1,151) (x3,221) (x5,452) (x7,478) (x10,549)
        };
        
    \end{axis}
    
  \end{tikzpicture}}
    \resizebox{\linewidth}{!}{\begin{tikzpicture}
      \begin{customlegend}[legend entries={\lexicon, DeepSeek R1, OpenAI o3, Gemini 2.5 Pro, Claude 3.7 Sonnet, GPT-5}, legend columns=-1, legend style={draw=none}]
          \addlegendimage{yellow!60!black, area legend, fill}
          \addlegendimage{red!60!black, area legend, fill}
          \addlegendimage{blue!60!black, area legend, fill}
          \addlegendimage{green!60!black, area legend, fill}
          \addlegendimage{pink!60!black, area legend, fill}
          \addlegendimage{orange!60!black, area legend, fill}  
      \end{customlegend}
\end{tikzpicture}}
\caption{\textbf{Execution Time vs. number of constraints (average optimal cost)}. The vertical axes show execution time in seconds. 
Standard deviations are small, and thus omitted.
}
\label{fig:times}
\end{figure}
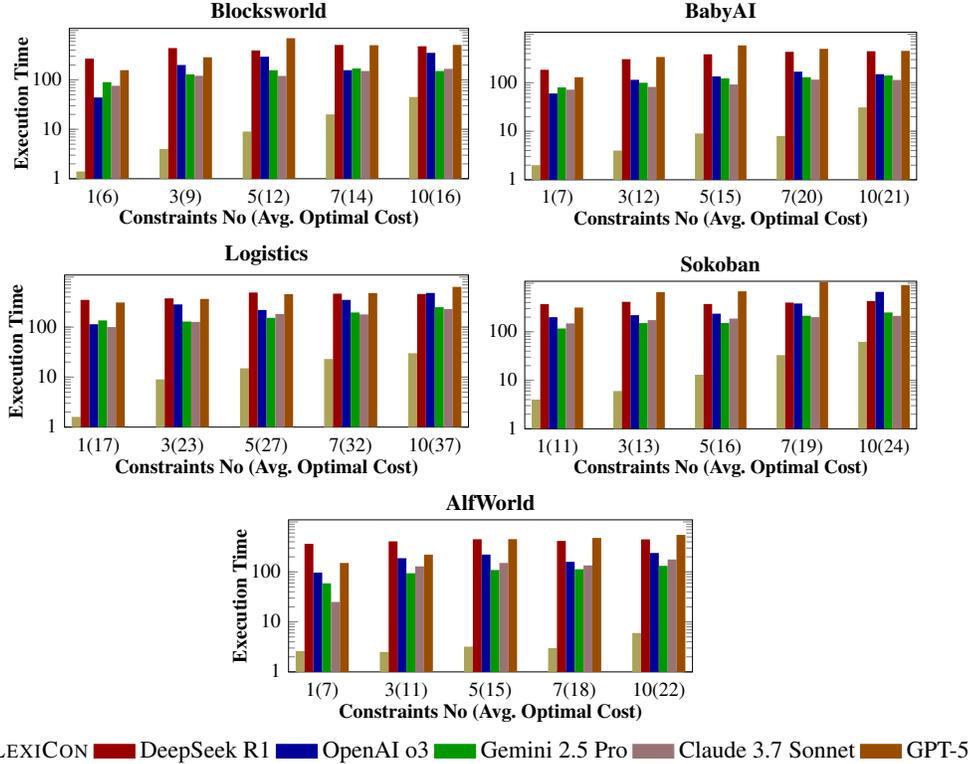

We evaluated 5 LLMs with thinking token generation capabilities, i.e., DeepSeek R1~\citep{deepseek_r1}, OpenAI o3~\citep{openai-o3}, Gemini-2.5 Pro~\citep{google2025gemini2.5}, Claude 3.7 Sonnet (Extended Thinking)~\citep{claude_37_sonnet}, and GPT-5~\cite{openai-gpt5}. We also tested 4 LLMs that that do not support thinking capabilities, i.e., GPT 4.1~\citep{openai2025gpt41}, DeepSeek V3~\cite{deepseekv3technical}, Claude 3.7 Sonnet (no extended thinking), and Gemini 2.0 Pro~\citep{gemini_20_flash}, on benchmarks generated by \lexicon. Each environment consisted of 150 problems, with the number of constraints in $\{1,3,5,7,10\}$.

\textbf{Can LLMs perform constrained planning?} Figure~\ref{fig:llm_results} displays our results. All data points were produced over 30 executions. For the sake of comparison, we also included performance measurements on unconstrained problems. \textbf{LLMs without explicit thinking typically failed to produce an optimal---or even valid---plan for problems involving more than one constraint}. For this class of models, we only report the best-performing configuration, which was GPT-4.1 with Chain-of-Thought (CoT) prompting~\cite{chain_of_thought}. For the performance of the remaining models, see Appendix~\ref{sec: additional results}.
In contrast, reasoning models frequently succeeded in producing optimal plans for problems with a few constraints. However, \textbf{the capabilities of reasoning LLMs deteriorated sharply as the number of constraints increased}. In the majority of problems including 10 constraints, most LLMs failed to even produce a suboptimal plan. 
In the case of o3, e.g., in Blockworld, the optimal planning accuracy over constraint range $\{1, 3, 5, 7, 10\}$ was [76\%, 30\%, 26\%, 10\%, 0].

Interestingly, as shown in Figure~\ref{fig:benchmark_stats}, we observe that the \textbf{number of thinking tokens increases with the optimal plan length} (i.e., optimal cost), suggesting that the reasoning models engage in deeper reasoning when the task demands it. However, as highlighted in Figure \ref{fig:llms_on_running_example}, their \emph{soundness} declines for longer plans, due to: invalid actions from precondition violations (e.g., attempting to pick up a box without facing it), hallucinated states (e.g., perceiving an empty space where a door exists), misinterpreted constraints, and loss of state tracking.

\textbf{Do LLMs show structured planning behaviour?} We qualitatively analysed the reasoning traces of R1 by annotating its ``thinking'' steps and mapping them to classical planning strategies. (The reasoning traces of the LLMs that performed better on our benchmark were not available to us.) Figure~\ref{fig:r1-annotated-trace} shows a representative example. We observed the following behaviours:
\begin{compactitem}
    \item \textbf{Bilevel Planning}. R1 decomposes the goal into high-level subgoals and performs subgoal planning.
    \item \textbf{Backtracking}. R1 can backtrack and generate a more optimized subgoal plan (e.g, instead of having to go back to pick an object, carry it with you).
    \item \textbf{Lookahead Search}. R1 generates multiple rollout paths and selects the optimal action.
    \item \textbf{Self-Reflection}. R1 frequently re-evaluates the state and the selected actions, checking constraint satisfaction and exploring alternatives, towards optimising the plan.
\end{compactitem}

Despite these interesting behaviours, \textbf{R1 does not show the structured search behaviour necessary for optimal planning}, like maintaining different search paths simultaneously. Instead, it tries to generate a valid plan that satisfies all constraints, and subsequently attempts to shorten that plan, towards finding an optimal one; this is not guaranteed to work for arbitrary planning problems.


%
\textbf{Can \lexicon enable real-time evaluation of LLM Planners?} A key advantage of \lexicon\ is its ability to generate arbitrary constrained planning problems on demand. This allows for on-the-fly evaluation of LLMs on problems of varying complexity, rather than relying on static, offline benchmarks. In our setup, \lexicon generates a new problem while the LLM is still solving the previous one, enabling a seamless and adaptive evaluation pipeline.
To test this, we compared the average time \lexicon\ takes to generate and verify a problem (in natural language and PDDL) with the average time an LLM takes to solve it\footnote{Note that LLM solve time also depends on API latency, though \lexicon\ remains significantly faster.}. As shown in Figure~\ref{fig:times}, \textbf{\lexicon\ is roughly one order of magnitude faster than LLM-based planning, making real-time evaluation feasible}.

\section{Summary \& Future Work}\label{sec:summary}

We proposed \lexicon, an extensible NL-based benchmark generator for planning under temporal constraints.
Our generator is able to produce task-aware constraints for an arbitrary planning problem and verify solutions suggested by LLMs at scale.
Our experiments showed that there is a limit of problem constrainedness that LLMs cannot cope with, even for models with reasoning capabilities. 
We aim to extend \lexicon with partially-observable environments and uncertain observations, as well as a wider class of constraints, including constraints on actions and on continuous states, paving the way for evaluating language-based agents on real reinforcement learning settings.

\section{Limitations}
\label{limitations}
Our simulator does not support parallel episode execution, unlike standard RL environments such as MuJoCo~\citep{mujoco} or Atari~\citep{atari-arcade}, which can be parallelized using tools like \texttt{AsyncVectorEnv} (Gymnasium)~\citep{gymnasium} or \texttt{SubprocVecEnv} (Stable-Baselines3)~\citep{stable-baselines}. In our case, multiprocessing is fully utilized for backend tasks such as generating feasible episodes. Furthermore, episode generation is significantly slower ($[1,100]$ s) due to the complexity of constraint satisfaction and simulation, limiting scalability compared to environments that support fast, parallel rollouts. 


\begin{ack}
    This work was supported by the Wallenberg AI, Autonomous Systems and Software Program (WASP) funded by the Knut and Alice Wallenberg Foundation.
    %
    This work was also supported by the Research Foundation - Flanders (FWO) under contract no G097720N. 
    %
\end{ack}


\bibliography{refs}




\section*{NeurIPS Paper Checklist}

\begin{enumerate}

\item {\bf Claims}
    \item[] Question: Do the main claims made in the abstract and introduction accurately reflect the paper's contributions and scope?
    \item[] Answer: \answerYes{} 
    \item[] Justification: \lexicon's reasoning engine (including the plan generator and verifier) are described in Section~\ref{sec:lexicon}. We also show how \lexicon can be easily extended to other environments in Section~\ref{sec:architecture}. Through our experiments on a range of environments (Section~\ref{subsection: evaluation setup}) and LLMs, we show that LLM including that of reasoning models struggle with constrained planning (Section~\ref{subsection: evaluation results}). Furthermore, as part of this submission, we provide a dataset for evaluation generated from \lexicon, along with the environment source files. 
    \item[] Guidelines:
    \begin{itemize}
        \item The answer NA means that the abstract and introduction do not include the claims made in the paper.
        \item The abstract and/or introduction should clearly state the claims made, including the contributions made in the paper and important assumptions and limitations. A No or NA answer to this question will not be perceived well by the reviewers. 
        \item The claims made should match theoretical and experimental results, and reflect how much the results can be expected to generalize to other settings. 
        \item It is fine to include aspirational goals as motivation as long as it is clear that these goals are not attained by the paper. 
    \end{itemize}

\item {\bf Limitations}
    \item[] Question: Does the paper discuss the limitations of the work performed by the authors?
    \item[] Answer: \answerYes{} 
    \item[] Justification: Section~\ref{limitations} 
    \item[] Guidelines:
    \begin{itemize}
        \item The answer NA means that the paper has no limitation while the answer No means that the paper has limitations, but those are not discussed in the paper. 
        \item The authors are encouraged to create a separate "Limitations" section in their paper.
        \item The paper should point out any strong assumptions and how robust the results are to violations of these assumptions (e.g., independence assumptions, noiseless settings, model well-specification, asymptotic approximations only holding locally). The authors should reflect on how these assumptions might be violated in practice and what the implications would be.
        \item The authors should reflect on the scope of the claims made, e.g., if the approach was only tested on a few datasets or with a few runs. In general, empirical results often depend on implicit assumptions, which should be articulated.
        \item The authors should reflect on the factors that influence the performance of the approach. For example, a facial recognition algorithm may perform poorly when image resolution is low or images are taken in low lighting. Or a speech-to-text system might not be used reliably to provide closed captions for online lectures because it fails to handle technical jargon.
        \item The authors should discuss the computational efficiency of the proposed algorithms and how they scale with dataset size.
        \item If applicable, the authors should discuss possible limitations of their approach to address problems of privacy and fairness.
        \item While the authors might fear that complete honesty about limitations might be used by reviewers as grounds for rejection, a worse outcome might be that reviewers discover limitations that aren't acknowledged in the paper. The authors should use their best judgment and recognize that individual actions in favor of transparency play an important role in developing norms that preserve the integrity of the community. Reviewers will be specifically instructed to not penalize honesty concerning limitations.
    \end{itemize}

\item {\bf Theory assumptions and proofs}
    \item[] Question: For each theoretical result, does the paper provide the full set of assumptions and a complete (and correct) proof?
    \item[] Answer: \answerNA{} 
    \item[] Justification: We do not any theoretical claims in the paper.
    \item[] Guidelines:
    \begin{itemize}
        \item The answer NA means that the paper does not include theoretical results. 
        \item All the theorems, formulas, and proofs in the paper should be numbered and cross-referenced.
        \item All assumptions should be clearly stated or referenced in the statement of any theorems.
        \item The proofs can either appear in the main paper or the supplemental material, but if they appear in the supplemental material, the authors are encouraged to provide a short proof sketch to provide intuition. 
        \item Inversely, any informal proof provided in the core of the paper should be complemented by formal proofs provided in appendix or supplemental material.
        \item Theorems and Lemmas that the proof relies upon should be properly referenced. 
    \end{itemize}

    \item {\bf Experimental result reproducibility}
    \item[] Question: Does the paper fully disclose all the information needed to reproduce the main experimental results of the paper to the extent that it affects the main claims and/or conclusions of the paper (regardless of whether the code and data are provided or not)?
    \item[] Answer: \answerYes{} 
    \item[] Justification: We provide the data and the code we used in our submission, and reproducibility instructions in the Appendix~\ref{appendix: reproducibility}. We also provide the LLM parameters in Appendix~\ref{appendix: llm parameters}.
    \item[] Guidelines: 
    \begin{itemize}
        \item The answer NA means that the paper does not include experiments.
        \item If the paper includes experiments, a No answer to this question will not be perceived well by the reviewers: Making the paper reproducible is important, regardless of whether the code and data are provided or not.
        \item If the contribution is a dataset and/or model, the authors should describe the steps taken to make their results reproducible or verifiable. 
        \item Depending on the contribution, reproducibility can be accomplished in various ways. For example, if the contribution is a novel architecture, describing the architecture fully might suffice, or if the contribution is a specific model and empirical evaluation, it may be necessary to either make it possible for others to replicate the model with the same dataset, or provide access to the model. In general. releasing code and data is often one good way to accomplish this, but reproducibility can also be provided via detailed instructions for how to replicate the results, access to a hosted model (e.g., in the case of a large language model), releasing of a model checkpoint, or other means that are appropriate to the research performed.
        \item While NeurIPS does not require releasing code, the conference does require all submissions to provide some reasonable avenue for reproducibility, which may depend on the nature of the contribution. For example
        \begin{enumerate}
            \item If the contribution is primarily a new algorithm, the paper should make it clear how to reproduce that algorithm.
            \item If the contribution is primarily a new model architecture, the paper should describe the architecture clearly and fully.
            \item If the contribution is a new model (e.g., a large language model), then there should either be a way to access this model for reproducing the results or a way to reproduce the model (e.g., with an open-source dataset or instructions for how to construct the dataset).
            \item We recognize that reproducibility may be tricky in some cases, in which case authors are welcome to describe the particular way they provide for reproducibility. In the case of closed-source models, it may be that access to the model is limited in some way (e.g., to registered users), but it should be possible for other researchers to have some path to reproducing or verifying the results.
        \end{enumerate}
    \end{itemize}

\item {\bf Open access to data and code}
    \item[] Question: Does the paper provide open access to the data and code, with sufficient instructions to faithfully reproduce the main experimental results, as described in supplemental material?
    \item[] Answer: \answerYes{} 
    \item[] Justification: We provide open access data and code URLs in our submission. Reproducibility instructions are provided in the Appendix~\ref{appendix: reproducibility}.
    \item[] Guidelines:
    \begin{itemize}
        \item The answer NA means that paper does not include experiments requiring code.
        \item Please see the NeurIPS code and data submission guidelines (\url{https://nips.cc/public/guides/CodeSubmissionPolicy}) for more details.
        \item While we encourage the release of code and data, we understand that this might not be possible, so “No” is an acceptable answer. Papers cannot be rejected simply for not including code, unless this is central to the contribution (e.g., for a new open-source benchmark).
        \item The instructions should contain the exact command and environment needed to run to reproduce the results. See the NeurIPS code and data submission guidelines (\url{https://nips.cc/public/guides/CodeSubmissionPolicy}) for more details.
        \item The authors should provide instructions on data access and preparation, including how to access the raw data, preprocessed data, intermediate data, and generated data, etc.
        \item The authors should provide scripts to reproduce all experimental results for the new proposed method and baselines. If only a subset of experiments are reproducible, they should state which ones are omitted from the script and why.
        \item At submission time, to preserve anonymity, the authors should release anonymized versions (if applicable).
        \item Providing as much information as possible in supplemental material (appended to the paper) is recommended, but including URLs to data and code is permitted.
    \end{itemize}

\item {\bf Experimental setting/details}
    \item[] Question: Does the paper specify all the training and test details (e.g., data splits, hyperparameters, how they were chosen, type of optimizer, etc.) necessary to understand the results?
    \item[] Answer: \answerYes{} 
    \item[] Justification: We provide the parameters, e.g., number of constraints, we used for problem generation. The specifics of the planning domains used, the technical details of the simulation, and the LLM hyperparameters used for planning are described in the Appendix~\ref{sec:appendix_pddl},~\ref{appendix: reproducibility}.
    \item[] Guidelines:
    \begin{itemize}
        \item The answer NA means that the paper does not include experiments.
        \item The experimental setting should be presented in the core of the paper to a level of detail that is necessary to appreciate the results and make sense of them.
        \item The full details can be provided either with the code, in appendix, or as supplemental material.
    \end{itemize}

\item {\bf Experiment statistical significance}
    \item[] Question: Does the paper report error bars suitably and correctly defined or other appropriate information about the statistical significance of the experiments?
    \item[] Answer: \answerYes{} 
    \item[] Justification: Error bars do not apply in Figures \ref{fig:llm_results} and \ref{fig:benchmark_stats} as we do not present average values of individual runs. For Figure \ref{fig:benchmark_stats}, we show statistical significance via Pearson correlations. In Figure \ref{fig:times}, we omit error bars because they are small.  
    \item[] Guidelines:
    \begin{itemize}
        \item The answer NA means that the paper does not include experiments.
        \item The authors should answer "Yes" if the results are accompanied by error bars, confidence intervals, or statistical significance tests, at least for the experiments that support the main claims of the paper.
        \item The factors of variability that the error bars are capturing should be clearly stated (for example, train/test split, initialization, random drawing of some parameter, or overall run with given experimental conditions).
        \item The method for calculating the error bars should be explained (closed form formula, call to a library function, bootstrap, etc.)
        \item The assumptions made should be given (e.g., Normally distributed errors).
        \item It should be clear whether the error bar is the standard deviation or the standard error of the mean.
        \item It is OK to report 1-sigma error bars, but one should state it. The authors should preferably report a 2-sigma error bar than state that they have a 96\% CI, if the hypothesis of Normality of errors is not verified.
        \item For asymmetric distributions, the authors should be careful not to show in tables or figures symmetric error bars that would yield results that are out of range (e.g. negative error rates).
        \item If error bars are reported in tables or plots, The authors should explain in the text how they were calculated and reference the corresponding figures or tables in the text.
    \end{itemize}

\item {\bf Experiments compute resources}
    \item[] Question: For each experiment, does the paper provide sufficient information on the computer resources (type of compute workers, memory, time of execution) needed to reproduce the experiments?
    \item[] Answer: \answerYes{} 
    \item[] Justification: We provide the specifications of the PC that runs our simulator (Section~\ref{subsection: evaluation setup}). The LLMs are run through their APIs with their maximum allowed token limits. Execution times are recorded (Figure~\ref{fig:times}).
    \item[] Guidelines:
    \begin{itemize}
        \item The answer NA means that the paper does not include experiments.
        \item The paper should indicate the type of compute workers CPU or GPU, internal cluster, or cloud provider, including relevant memory and storage.
        \item The paper should provide the amount of compute required for each of the individual experimental runs as well as estimate the total compute. 
        \item The paper should disclose whether the full research project required more compute than the experiments reported in the paper (e.g., preliminary or failed experiments that didn't make it into the paper). 
    \end{itemize}
    
\item {\bf Code of ethics}
    \item[] Question: Does the research conducted in the paper conform, in every respect, with the NeurIPS Code of Ethics \url{https://neurips.cc/public/EthicsGuidelines}?
    \item[] Answer: \answerYes{} 
    \item[] Justification: No ethics issues.
    \item[] Guidelines:
    \begin{itemize}
        \item The answer NA means that the authors have not reviewed the NeurIPS Code of Ethics.
        \item If the authors answer No, they should explain the special circumstances that require a deviation from the Code of Ethics.
        \item The authors should make sure to preserve anonymity (e.g., if there is a special consideration due to laws or regulations in their jurisdiction).
    \end{itemize}

\item {\bf Broader impacts}
    \item[] Question: Does the paper discuss both potential positive societal impacts and negative societal impacts of the work performed?
    \item[] Answer: \answerNA{} 
    \item[] Justification: While there is no direct societal impact, if constrained planning is see from a safety perspective, our work can be seen as way to evaluate whether LLM-based agents can adhere to safety constraints.
    \item[] Guidelines:
    \begin{itemize}
        \item The answer NA means that there is no societal impact of the work performed.
        \item If the authors answer NA or No, they should explain why their work has no societal impact or why the paper does not address societal impact.
        \item Examples of negative societal impacts include potential malicious or unintended uses (e.g., disinformation, generating fake profiles, surveillance), fairness considerations (e.g., deployment of technologies that could make decisions that unfairly impact specific groups), privacy considerations, and security considerations.
        \item The conference expects that many papers will be foundational research and not tied to particular applications, let alone deployments. However, if there is a direct path to any negative applications, the authors should point it out. For example, it is legitimate to point out that an improvement in the quality of generative models could be used to generate deepfakes for disinformation. On the other hand, it is not needed to point out that a generic algorithm for optimizing neural networks could enable people to train models that generate Deepfakes faster.
        \item The authors should consider possible harms that could arise when the technology is being used as intended and functioning correctly, harms that could arise when the technology is being used as intended but gives incorrect results, and harms following from (intentional or unintentional) misuse of the technology.
        \item If there are negative societal impacts, the authors could also discuss possible mitigation strategies (e.g., gated release of models, providing defenses in addition to attacks, mechanisms for monitoring misuse, mechanisms to monitor how a system learns from feedback over time, improving the efficiency and accessibility of ML).
    \end{itemize}
    
\item {\bf Safeguards}
    \item[] Question: Does the paper describe safeguards that have been put in place for responsible release of data or models that have a high risk for misuse (e.g., pretrained language models, image generators, or scraped datasets)?
    \item[] Answer: \answerNA{} 
    \item[] Justification: We use simulators from Classical AI domains that are publicly available. These are simple toy-based setups for evaluating decision-making agents.
    \item[] Guidelines:
    \begin{itemize}
        \item The answer NA means that the paper poses no such risks.
        \item Released models that have a high risk for misuse or dual-use should be released with necessary safeguards to allow for controlled use of the model, for example by requiring that users adhere to usage guidelines or restrictions to access the model or implementing safety filters. 
        \item Datasets that have been scraped from the Internet could pose safety risks. The authors should describe how they avoided releasing unsafe images.
        \item We recognize that providing effective safeguards is challenging, and many papers do not require this, but we encourage authors to take this into account and make a best faith effort.
    \end{itemize}

\item {\bf Licenses for existing assets}
    \item[] Question: Are the creators or original owners of assets (e.g., code, data, models), used in the paper, properly credited and are the license and terms of use explicitly mentioned and properly respected?
    \item[] Answer: \answerYes{} 
    \item[] Justification: We provide citations for the planning domains we employed and the models we evaluated. The licenses of the modules used as components in our simulator, like the off-the-self planner, permit our usage.
    \item[] Guidelines:
    \begin{itemize}
        \item The answer NA means that the paper does not use existing assets.
        \item The authors should cite the original paper that produced the code package or dataset.
        \item The authors should state which version of the asset is used and, if possible, include a URL.
        \item The name of the license (e.g., CC-BY 4.0) should be included for each asset.
        \item For scraped data from a particular source (e.g., website), the copyright and terms of service of that source should be provided.
        \item If assets are released, the license, copyright information, and terms of use in the package should be provided. For popular datasets, \url{paperswithcode.com/datasets} has curated licenses for some datasets. Their licensing guide can help determine the license of a dataset.
        \item For existing datasets that are re-packaged, both the original license and the license of the derived asset (if it has changed) should be provided.
        \item If this information is not available online, the authors are encouraged to reach out to the asset's creators.
    \end{itemize}

\item {\bf New assets}
    \item[] Question: Are new assets introduced in the paper well documented and is the documentation provided alongside the assets?
    \item[] Answer: \answerYes{} 
    \item[] Justification: We provide documented code with execution scripts and examples in our code submission, and a technical description of our system in the technical appendix (Appendix~\ref{sec:appendix_pddl}).
    \item[] Guidelines:
    \begin{itemize}
        \item The answer NA means that the paper does not release new assets.
        \item Researchers should communicate the details of the dataset/code/model as part of their submissions via structured templates. This includes details about training, license, limitations, etc. 
        \item The paper should discuss whether and how consent was obtained from people whose asset is used.
        \item At submission time, remember to anonymize your assets (if applicable). You can either create an anonymized URL or include an anonymized zip file.
    \end{itemize}

\item {\bf Crowdsourcing and research with human subjects}
    \item[] Question: For crowdsourcing experiments and research with human subjects, does the paper include the full text of instructions given to participants and screenshots, if applicable, as well as details about compensation (if any)? 
    \item[] Answer: \answerNA{} 
    \item[] Justification: No crowdsourcing or research with human subjects. 
    \item[] Guidelines:
    \begin{itemize}
        \item The answer NA means that the paper does not involve crowdsourcing nor research with human subjects.
        \item Including this information in the supplemental material is fine, but if the main contribution of the paper involves human subjects, then as much detail as possible should be included in the main paper. 
        \item According to the NeurIPS Code of Ethics, workers involved in data collection, curation, or other labor should be paid at least the minimum wage in the country of the data collector. 
    \end{itemize}

\item {\bf Institutional review board (IRB) approvals or equivalent for research with human subjects}
    \item[] Question: Does the paper describe potential risks incurred by study participants, whether such risks were disclosed to the subjects, and whether Institutional Review Board (IRB) approvals (or an equivalent approval/review based on the requirements of your country or institution) were obtained?
    \item[] Answer: \answerNA{} 
    \item[] Justification: No crowdsourcing or research with human subjects. 
    \item[] Guidelines:
    \begin{itemize}
        \item The answer NA means that the paper does not involve crowdsourcing nor research with human subjects.
        \item Depending on the country in which research is conducted, IRB approval (or equivalent) may be required for any human subjects research. If you obtained IRB approval, you should clearly state this in the paper. 
        \item We recognize that the procedures for this may vary significantly between institutions and locations, and we expect authors to adhere to the NeurIPS Code of Ethics and the guidelines for their institution. 
        \item For initial submissions, do not include any information that would break anonymity (if applicable), such as the institution conducting the review.
    \end{itemize}

\item {\bf Declaration of LLM usage}
    \item[] Question: Does the paper describe the usage of LLMs if it is an important, original, or non-standard component of the core methods in this research? Note that if the LLM is used only for writing, editing, or formatting purposes and does not impact the core methodology, scientific rigorousness, or originality of the research, declaration is not required.
    \item[] Answer: \answerYes{} 
    \item[] Justification: While our simulator does not use LLMs, we evaluate LLM planning abilities using episodes/data generated from our simulator. All the LLMs have been adequately cited (Section~\ref{subsection: evaluation setup}), their evaluation results discussed (Section~\ref{subsection: evaluation results}), and their generation parameters provide for reproducibility (Appendix~\ref{appendix: llm parameters})
    \item[] Guidelines:
    \begin{itemize}
        \item The answer NA means that the core method development in this research does not involve LLMs as any important, original, or non-standard components.
        \item Please refer to our LLM policy (\url{https://neurips.cc/Conferences/2025/LLM}) for what should or should not be described.
    \end{itemize}

\end{enumerate}

\newpage
\appendix

\begin{center}

\section*{\Large Appendix}

\end{center}

This document contains supplementary material for our paper, along with code execution and reproducibility instructions for our experiments.
Its structure is the following.
Appendix \ref{sec:planning problem} describes \lexicon's reasoning engine.
Appendix \ref{sec: additional results} exemplifies indicative errors in LLM planning, while also highlighting some experimental results that were omitted from the paper due to space limitations.
Appendix \ref{appendix: reproducibility} provides the hyperparameters set for each LLM in our experiments, along with the steps for running our code and reproducing our experiments.

\section{Reasoning Engine}\label{sec:planning problem}
First, we specify the class of planning problems that may be generated and have candidate solutions verified by \lexicon's reasoning engine.
Subsequently, we describe its two main modules: the constraint planning problem generator and the automated LLM plan verifier.
\subsection{Class of Planning Problems in \lexicon}\label{sec:appendix_pddl}

\lexicon\ may generate and verify candidate solutions for planning domains expressed in a PDDL fragment that includes the following syntactic components.
\begin{itemize}
    \item Basic STRIPS, i.e., actions with conjunctive preconditions, and atom addition and deletion effects~\cite{DBLP:journals/ai/FikesN71}.
    \item ADL, i.e., equalities, actions with negated, disjunctive and quantified preconditions, as well as conditional and universally quantified effects~\cite{DBLP:conf/kr/Pednault89}.
    \item The qualitative state-trajectory constraints found in PDDL3.0~\cite{DBLP:journals/ai/GereviniHLSD09}.
\end{itemize}
We formulate this fragment of PDDL, loosely following~\cite{DBLP:conf/aips/BonassiGPS21} for the notation, and using the term ``constraint'' to refer to a qualitative state-trajectory constraint of PDDL3.0 for brevity.

A constrained planning problem is a tuple $\pddlProblem\val (\pddlAtoms, \pddlActions, \pddlInit, \pddlGoal, \pddlConstraints)$, where $\pddlAtoms$ is a set of atoms, $\pddlActions$ is a set of actions, $\pddlInit\subseteq \pddlAtoms$ is an initial state, $\pddlGoal$ is a formula over $\pddlAtoms$ denoting the goal of the problem, and $\pddlConstraints$ is a set of constraints.
Each action $a\in\pddlActions$ comprises a precondition $\actionPrec{a}$, which is a formula over $\pddlAtoms$, and a set of conditional effects $\actionEff{a}$.
Each conditional effect in $\actionEff{a}$ is an expression $c\rhd e$, where $c$ is a formula and $e$ is a set of literals, both constructed based on the atoms in $\pddlAtoms$.
We use $e^+$ (resp.~$e^-$) to denote the positive (negative) literals in $e$.
A state $s\subseteq\pddlAtoms$ contains the atoms that are true in $s$.
An action $a$ is applicable in state $s$ if $s\models \actionPrec{a}$, and its application yields state $s'\val (s\setminus \bigcup_{c\rhd e\in \actionEff{a}: s\models c} e^-)\cup \bigcup_{c\rhd e\in \actionEff{a}: s\models c} e^+$, which we often denote with $s'\val s[a]$.

\lexicon\ supports the following types of constraints: $\Always$, $\Sometime$, $\AtMostOnce$, $\SometimeBefore$ and $\SometimeAfter$.
Considering grounded formulas $\phi$ and $\psi$ over $\pddlAtoms$ in negation normal form, and a sequence of states $\stateseq$ over $\pddlAtoms$, these constraint types are defined as follows:
\begin{itemize}
    \item $\stateseq\models\Always(\phi)$ (or $\alw{\phi}$) iff $\forall s\in\stateseq$: $s\models\phi$.
    \item $\stateseq\models\Sometime(\phi)$ ($\st{\phi}$) iff $\exists s\in\stateseq$: $s\models\phi$.
    \item $\stateseq\models\AtMostOnce(\phi)$ ($\amo{\phi}$) iff $\phi$ is true in at most one continuous subsequence of $\stateseq$.
    \item $\stateseq\models\SometimeBefore(\phi, \psi)$ ($\stb{\phi}{\psi}$) requires that, if $\exists s\in\stateseq: s\models\phi$, then there is a state $s'$ before $s$ in $\stateseq$, such that $s'\models\psi$.
    \item $\stateseq\models\SometimeAfter(\phi, \psi)$ ($\sta{\phi}{\psi}$) requires that, if $\exists s\in\stateseq: s\models\phi$, then $s\models\psi$ or there is a state $s'$ after $s$ in $\stateseq$ such that $s'\models\psi$.
\end{itemize}

Given a constrained planning problem $\pddlProblem\val (\pddlAtoms, \pddlActions, \pddlInit, \pddlGoal, \pddlConstraints)$, a plan $\plan$ for $\pddlProblem$ is a sequence of actions $(a_0, \dots, a_{n\minus 1})$ from  set $\pddlActions$.
$\plan$ is a valid plan for $\pddlProblem$ iff there exists a sequence of states $\sigma\val (s_0, \dots, s_n)$ such that $s_0\val I$, $\forall i\in \{0, \dots, n\minus 1\}$ we have $s_i\models \actionPrec{a_i}$ and $s_{i\plus 1}\val s_i[a_i]$, $s_n\models \pddlGoal$, and $\forall \constr\in\pddlConstraints$ we have $\sigma\models\constr$.
We define the cost of a plan as the number of actions it includes.
An optimal plan $\planopt$ for a problem $\pddlProblem$ is a valid plan whose cost is minimal among all valid plans for $\pddlProblem$, i.e., there is no valid plan for $\pddlProblem$ that has a lower cost than $\planopt$.

\subsection{Constrained Planning Problem Generator}

We focus on the ``Constraint Generator'' of \lexicon, as the remaining modules of our reasoning engine that are used for problem generation are off-the-shelf planners and compilers (see Figure \ref{fig:lexicon}).
Our constraint generator receives as input an unconstrained PDDL problem $\pddlProblemUnc\val (\pddlAtoms, \pddlActions, \pddlInit, \pddlGoal)$ and an optimal plan $\planopt$ for $\pddlProblemUnc$, and outputs a constrained PDDL problem $\pddlProblem\val (\pddlAtoms, \pddlActions, \pddlInit, \pddlGoal, \pddlConstraints)$. 
The challenge here is to construct constraint set $\pddlConstraints$ in an informed manner, considering problem $\pddlProblemUnc$ and plan $\planopt$.
In particular, we may add a constraint $\constr$ in $\pddlConstraints$ only if $\constr$ is a meaningful constraint given $\pddlProblemUnc$ and $\planopt$, i.e., the inclusion of $\constr$ makes $\planopt$ an invalid plan for $\pddlProblemUnc$, potentially complicating the planning problem, while maintaining problem solvability and being non-redundant with respect to the constraints that were previously added in $\pddlConstraints$.

In order to produce such a meaningful constraint $\constr$, we proceed as follows.
\begin{enumerate}
\item We identify a set of conditions under which a literal is not suitable for inclusion in $\constr$, in the sense that its inclusion potentially results in $\constr$ not being meaningful for the problem. 
%
\item We sample literals that do not satisfy the conditions identified in the previous step, and consider whether they should be included in $\constr$.
For each sampled literal $l$, we verify that it is consistent with, and not subsumed by, the literals that were previously added in $\constr$, taking into account a (possibly empty) set of domain axioms.
If this is the case, then we add $l$ in $\constr$.
We continue this process until $\constr$ has reached a specified degree of compositionality, which may be controlled by the user.
\item We verify that the generated constraint $\constr$ is consistent with, and not subsumed by, the constraints that were previously added in $\pddlConstraints$, in which case we add $\constr$ in $\pddlConstraints$.
\end{enumerate}
We continue this process until the size of $\pddlConstraints$ has reached the number of constraints requested by the user.

\begin{algorithm}
\caption{$\Always$ Constraint Generator}\label{alg:always_constraint_sampler}
\begin{algorithmic}[1]
\Require State changes $\stateseq$ induced by executing plan $\planopt$, unconstrained problem $\pddlProblemUnc\val (\pddlAtoms, \pddlActions, \pddlInit, \pddlGoal)$, constraint set so far $\pddlConstraints$, domain axioms $\domainformulas$, possible user parameter values $\userparams$
\Ensure New constraint set $\pddlConstraints\cup \{\constr\}$
\State $\constrop, \literalsno \gets sample\_parameters(\userparams)$,\ $literals\gets\emptyset$ \label{line:sample_constraint_params}
\For{$\literalsno$ iterations} \label{line:literals_sample_loop}
    \State $l \gets sample\_literal(\pddlAtoms)$ \label{line:sample_literal}
    \If{$l\rightarrow \pddlGoal$ \textbf{or} $\pddlGoal\rightarrow \neg l$ \textbf{or} \textbf{not} $(\pddlInit \models l)$ \textbf{or} $\forall s\in\stateseq: s\models l$}~\textbf{goto} \ref{line:sample_literal}  \label{line:skip_literal1}
    \EndIf
    \For{$l'$ \textbf{in} $literals$} \label{line:literal_compatible_loop}
    \If{$l\val l'$ \textbf{or} ($\constrop\val\wedge$ \textbf{and} $\domainformulas\models \neg (l \wedge l')$)}~\textbf{goto} \ref{line:sample_literal} \label{line:skip_literal2}
    \EndIf
    \EndFor
    \State $literals.\append(l)$ \label{line:appendl}
\EndFor
\State \textbf{if} $\constrop\val \wedge$ \textbf{then} $\phi\gets \bigwedge_{l\in literals} l$ \textbf{else} $\phi\gets \bigvee_{l\in literals} l$ \label{line:phi}
\State $\constr\gets \alw{\phi}$ \label{line:q}
\For{$\constr'\in\pddlConstraints$} \label{line:for_constraints}
    \If{$\constr'\val \alw{\phi'}$ \textbf{and} ($\domainformulas\models (\phi\rightarrow \phi')$ \textbf{or} $\domainformulas\models (\phi'\rightarrow \phi)$ \textbf{or} $\domainformulas\models \neg (\phi \wedge \phi')$)}~\textbf{goto} \ref{line:sample_constraint_params} \label{line:other_always}
    \ElsIf{$\constr'\val \st{\phi'}$ \textbf{and} ($\domainformulas\models (\phi\rightarrow \phi')$ \textbf{or} $\domainformulas\models \neg (\phi \wedge \phi')$)}~\textbf{goto} \ref{line:sample_constraint_params} \label{line:other_sometime}
    \ElsIf{$\constr'\val \amo{\phi'}$ \textbf{and} ($\domainformulas\models (\phi\rightarrow \phi')$ \textbf{or} $\domainformulas\models \neg (\phi \wedge \phi')$)}~\textbf{goto} \ref{line:sample_constraint_params} \label{line:other_atmostonce}
    \ElsIf{$\constr'\val \stb{\phi'}{\psi'}$ \textbf{and} ($\domainformulas\models (\phi\rightarrow \phi')$ \textbf{or} $\domainformulas\models (\phi\rightarrow \psi')$ \label{line:other_stb_1} \\ 
    \qquad\qquad\qquad\qquad\qquad\qquad\quad\textbf{or} $\domainformulas\models \neg (\phi \wedge \phi')$  \textbf{or} $\domainformulas\models \neg (\phi \wedge \psi')$)}~\textbf{goto} \ref{line:sample_constraint_params} \label{line:other_stb_2}
    \ElsIf{$\constr'\val \sta{\phi'}{\psi'}$ \textbf{and} ($\domainformulas\models (\phi\rightarrow \psi')$ \textbf{or} $\domainformulas\models \neg (\phi \wedge \phi')$  \textbf{or} $\domainformulas\models \neg (\phi \wedge \psi')$)} \label{line:other_sta_1}
        \State \textbf{goto} \ref{line:sample_constraint_params} \label{line:other_sta_2}
    \EndIf
\EndFor
\State \textbf{return} $\pddlConstraints\cup \{\constr\}$ \label{line:return}
\end{algorithmic}
\end{algorithm} 

The above procedure for generating a constraint is adapted for each possible type of constraint.
Constraint consistency and subsumption, e.g., is defined differently for each constraint type.
As an example, Algorithm \ref{alg:always_constraint_sampler} outlines the procedure for constructing an $\Always$ constraint $\alw{\phi}$.
We start by sampling a Boolean operation $\constrop$ and a number of literals $\literalsno$ for $\phi$, taking into account the parameters that are optionally provided by the user (see line \ref{line:sample_constraint_params} of Algorihtm \ref{alg:always_constraint_sampler}).
Subsequently, we generate $\literalsno$ literals that are suitable for inclusion in $\phi$ (lines \ref{line:literals_sample_loop}--\ref{line:appendl}).
We sample a literal $l$ based on the atoms of the problem $\pddlAtoms$ (line \ref{line:sample_literal}), and then evaluate a set of conditions such that, if $l$ satisfies one of them, then including $l$ in $\phi$ is not meaningful with respect to constraint $\alw{\phi}$.
For instance, it is not meaningful to include $l$ in $\alw{\phi}$ if (i) $l$ implies the goal $\pddlGoal$, as, in non-trivial problems where $\pddlInit\not\models\pddlGoal$, $l\rightarrow \pddlGoal$ implies that $l$ does not hold in the initial state $\pddlInit$, and thus cannot ``always'' hold; (ii) $\pddlGoal$ implies $\neg l$, because then $l$ cannot hold in the final state of a plan that brings about $\pddlGoal$; (iii) if $l$ does not hold in $\pddlInit$; or (iv) if $l$ is satisfied in every state in the sequence $\stateseq$ induced by executing plan $\planopt$, as, in that case, $\alw{l}$ is satisfied by optimal plan $\planopt$ of the unconstrained problem, and thus adding $l$ in $\alw{\phi}$ may not lead to a more complicated problem.
If any of the above conditions holds, then we drop $l$ and sample another literal for our constraint (line \ref{line:skip_literal1}).
Additionally, we resort to resampling if $l$ has already been added to $\constr$ in a previous step, or the selected operation $\constrop$ is a conjunction and $l$ is inconsistent with some other literal $l'$ in $\constr$, taking into account a (possibly empty) set of atemporal domain axioms (see lines \ref{line:literal_compatible_loop}--\ref{line:skip_literal2}).
If none of the above conditions is satisfied, then we add $l$ to the set of literals that will be used to construct $\constr$ (line \ref{line:appendl}).

After identifying $\literalsno$ literals that are suitable for constraint $\alw{\phi}$, we construct $\phi$ and $\alw{\phi}$ using the sampled operation $\constrop$ (see lines \ref{line:phi}--\ref{line:q} of Algorithm \ref{alg:always_constraint_sampler}).
Next, we need to verify whether $\alw{\phi}$ is inconsistent or redundant with respect to the constraints that are already present in $\pddlConstraints$.
To do this, for each constraint $\constr'$ in $\pddlConstraints$, we check if $\alw{\phi}$ is compatible with $\constr'$ (lines \ref{line:for_constraints}--\ref{line:other_sta_2}). 
If $\alw{\phi}$ is not compatible with some constraint in $\pddlConstraints$, then we drop $\alw{\phi}$ and generate another constraint.
Otherwise, if $\alw{\phi}$ is compatible with every constraint in $\pddlConstraints$, then we add it in $\pddlConstraints$ (line \ref{line:return}).
For example, $\constr\val \alw{\phi}$ is compatible with a $\Sometime$ constraint $\constr'\val \st{\phi'}$ if, according to the atemporal domain axioms, (i) $\phi$ does not imply $\phi'$, as $\phi\rightarrow \phi'$ would imply that constraint $\st{\phi'}$ is redundant given $\alw{\phi}$; and (ii) $\phi$ and $\phi'$ are consistent, because if they were inconsistent, it would be impossible to satisfy $\alw{\phi}$ given that $\st{\phi'}$ holds, i.e., in the case where $\phi'$ is true in at least one state of a valid plan.

\subsection{Automated LLM Plan Verifier}

\begin{algorithm}
\caption{LLM Plan Verifier}\label{alg:plan_verifier}
\begin{algorithmic}[1]
\Require LLM plan $\plannl$, compiled problem $\pddlProblemComp$, optimal cost $\costopt$
\Ensure Plan validation outcome: $\invalid$, $\suboptimal$ or $\optimal$
\State $s\gets I$ \label{line:stateinit}
\For{$\actionnl\in \plannl$} \label{line:actionnl}
\If{$pddl\_format(\actionnl)$}~$a\gets extract\_action(\actionnl)$ \label{line:extract_action}
\Else~$a\gets closest\_action\_edit\_distance(\actionnl, \pddlProblemUnc)$ \label{line:closest_action}
\EndIf
\State $s\gets simulate\_action(a, s, \pddlProblemUnc)$ \label{line:simulate_action}
\If{$s\val \mathsf{None}$}~\textbf{return} $\invalid$ \label{line:failed_action}
\EndIf
\EndFor
\If{$s\not\models\pddlGoal$}~\textbf{return} $\invalid$ \label{line:goal_not_reached}
\EndIf
\If{$length(\plannl) > \costopt$}~\textbf{return} $\suboptimal$ \label{line:suboptimal}
\EndIf
\State \textbf{return} $\optimal$ \label{line:optimal}
\end{algorithmic}
\end{algorithm} 

Algorithm \ref{alg:plan_verifier} outlines \lexicon's automated LLM plan verifier.
Its input is an LLM-generated plan $\plannl$, a PDDL problem $\pddlProblemComp$ that has been compiled with $\liftedtcore$ and the optimal cost $\costopt$ of $\pddlProblemComp$ that was discovered by \lexicon during constrained problem generation.
(Recall that a plan that is valid for the compiled version of a problem satisfies all the constraints in the original problem.)
Given this input, Algorithm \ref{alg:plan_verifier} reports whether $\plannl$ is an invalid plan, a valid but suboptimal plan, or a valid, optimal plan for problem $\pddlProblemComp$.
This is achieved in two steps: (i) simulating plan $\plannl$ over $\pddlProblemComp$ to verify its validity, and, if $\plannl$ is valid, (ii) comparing the length of $\plannl$ to the optimal cost $\costopt$ of $\pddlProblemComp$ in order to check whether $\plannl$ is optimal.

To initiate the simulation of plan $\plannl$ over problem $\pddlProblemComp$, we set variable $s$, tracking the state of the problem, to the initial state $\pddlInit$ of $\pddlProblemComp$ (line \ref{line:stateinit} of Algorithm \ref{alg:plan_verifier}), and iterate over the actions in $\plannl$, in order to sequentially simulate the effects of each one over $\pddlProblemComp$ (line \ref{line:actionnl}).
The prompt we use for LLM plan generation requests a specific format for LLM actions, so that they can be mapped directly to domain actiin in PDDL (line \ref{line:extract_action}).
In practice, however, LLM actions may deviate from this format; we handle such cases by mapping the LLM-generated action $a_{NL}$ to the PDDL domain action yielding the shortest edit distance from $a_{NL}$ (line \ref{line:closest_action}).
Both cases map $a_{NL}$ to a PDDL domain action $a$, which we apply on the current state $s$ of our plan simulation (line \ref{line:simulate_action}).
If the application of $a$ does not lead to a new state, then we deduce that either the preconditions of $a$ are not met in state $s$, or that the application of $a$ over state $s$ led to the violation of a constraint of the original problem.
Thus, in this case, we deduce that plan $\plannl$ is invalid.
If the simulation of all LLM-generated actions over $\pddlProblemComp$ succeeds, then we check whether the goal of the problem is satisfied in the final state $s$ reached in the simulation.
If the goal is not satisfied in $s$, then plan $\plannl$ is invalid (line \ref{line:goal_not_reached}).
Otherwise, if the goal is satisfied in $s$, then $\plannl$ is valid, and we proceed with checking whether $\plannl$ is optimal or not.
If the length of $\plannl$ is greater than $\costopt$, i.e., the optimal cost of $\pddlProblemComp$, then plan $\plannl$ is suboptimal (line \ref{line:suboptimal}).
Otherwise, the length of $\plannl$ is equal to $\costopt$, and thus $\plannl$ is an optimal plan for the problem (line \ref{line:optimal}).

\section{Additional Results}
\label{sec: additional results}

\subsection{Performance of Non-Reasoning LLMs}

\begin{table}[h]
\centering
\renewcommand{\arraystretch}{1.15} 
\begin{tabular}{lcccccccccc}
\textbf{Model} & \multicolumn{2}{c}{Blocksworld} & \multicolumn{2}{c}{BabyAI} & \multicolumn{2}{c}{Logistics} & \multicolumn{2}{c}{Sokoban} & \multicolumn{2}{c}{AlfWorld} \\
 & Opt. & Val. & Opt. & Val. & Opt. & Val. & Opt. & Val. & Opt. & Val. \\
\midrule
DeepSeek V3 & 0 & 7\% & 9\% & 17\% & 0 & 5\% & 0 & 0 & 15\% & 21\% \\
\makecell[l]{Claude 3.7 Sonnet\\ (no thinking)} & 0 & 0 & 12\% & 20\% & 5\% & 5\% & 0 & 0 & 3\% & 12\ \\
Gemini 2.0 Pro & 0 & 7\% & 0 & 0 & 0 & 0 & 0 & 0 & 0 & 0 \\
OpenAI GPT-4.1 & 3\% & 16\% & 17\% & 26\% & 0 & 3\% & 0 & 0 & 0 & 0 \\
\end{tabular}
\vspace{5pt}
\caption{Performance of non-reasoning LLMs on problems with one constraint. We measured the percentage of problems solved with an optimal plan (Opt.) and the percentage of problems solved with a valid, but possibly suboptimal, plan (Val.).}
\label{tab:non-reasoning}
\end{table}

We complement the experimental results in Figure \ref{fig:llm_results} of the paper with the performance of LLMs that do not use explicit thinking, i.e., DeepSeek V3, Claude 3.7 Sonnet (no extended thinking) and Gemini 2.0 Pro, on constrained planning problems.
Table \ref{tab:non-reasoning} displays their performance on each domain, in terms of plan optimality and plan validity, on problems that included one constraint.
None of these models was able to produce an optimal plan for a problem from our benchmark that included more than one constraint.

\subsection{LLM Action Format Compliance}

\begin{table}[h]
\centering
\renewcommand{\arraystretch}{1.15}
\resizebox{\linewidth}{!}{
\begin{tabular}{lccccccccccccccc}
\textbf{Model} & \multicolumn{3}{c}{Blocksworld} & \multicolumn{3}{c}{BabyAI} & \multicolumn{3}{c}{Logistics} & \multicolumn{3}{c}{Sokoban} & \multicolumn{3}{c}{AlfWorld} \\
 & 1 & 5 & 10 & 1 & 5 & 10 & 1 & 5 & 10 & 1 & 5 & 10 \\
\midrule
DeepSeek R1 & 0 & 50\% & 47\% & 6\% & 10\% & 23\% & 3\% & 7\% & 10\% & 3\% & 3\% & 3\% & 6\% & 0 & 0 \\
OpenAI o3 & 0 & 0 & 0 & 0 & 3\% & 3\% & 0 & 0 & 3\% & 0 & 0 & 3\% & 0 & 0 & 3\% \\
Gemini 2.5 Pro & 0 & 0 & 0 & 0 & 0 & 0 & 0 & 3\% & 3\% & 0 & 0 & 0 & 70\% & 78\% & 81\% \\
\makecell[l]{Claude 3.7 Sonnet\\ (with thinking)} & 27\% & 33\% & 47\% & 10\% & 12\% & 27\% & 33\% & 42\% & 60\% & 36\% & 45\% & 96\% & 15\% & 33\% & 39\% \\
GPT-5 & 3\% & 3\% & 3\% & 0 & 0 & 0 & 3\% & 0 & 0 & 0 & 0 & 0 & 0 & 3\% & 0 \\
\end{tabular}
}
\vspace{5pt}
\caption{Percentage of LLM-generated plans that could not be mapped directly into PDDL for problems with 1, 5 and 10 constraints.}
\label{tab:failed_parses}
\end{table}

During LLM plan verification, we measured the number of times an LLM-generated plan did not comply with the format we instructed the LLMs to follow via our prompt.
Table \ref{tab:failed_parses} displays our results on LLMs with reasoning capabilities, when instructed to suggest plans for problems with 1, 5 and 10 constraints.
Our results show that the responses of o3 and Gemini 2.5 Pro included, in almost all cases, a plan that conformed with the format of PDDL actions, and could thus be mapped directly into a PDDL plan, without needing to resort to distance calculations between LLM-generated actions and domain actions.
In contrast, the responses of R1 and Claude 3.7 Sonnet often deviated from the requested plan format, in which cases we needed to map some of the actions in the suggested plans into PDDL action via distance minimisation, in order to be able to verify these plans.


\section{Experimental Setup and Reproducibility}
\label{appendix: reproducibility}

First, we outline the hyperparameter values used in LLM executions.
Second, we provide a set of instructions for running \lexicon.
Third, we outline the steps for reproducing our experiments.

\subsection{Execution Parameters}
\label{appendix: llm parameters}

\begin{table}[h]
\centering
\begin{tabular}{lcc}
\textbf{Model} & max tokens & temperature \\
\midrule
DeepSeek R1 & 40K & 0.2 \\
OpenAI o3 & 100K & 1 \\
Gemini 2.5 Pro & 64K & 0.2 \\
Claude 3.7 Sonnet (with thinking) & 64K & 1 \\
GPT-5 & 128K & 0.2 \\
GPT-4.1 & 32K & 0.2 \\
DeepSeek V3 & 8K & 0.2 \\
Gemini 2.0 Pro & 8K & 0.2 \\
Claude 3.7 Sonnet (no thinking) & 64K & 0.2 \\
\end{tabular}
\vspace{5pt}
\caption{LLM hyperparameters.}
\label{tab:llm-hyperparams}
\end{table}

Table \ref{tab:llm-hyperparams} displays the values for the upper limit on generated tokens (including both completion and reasoning tokens) and the temperature hyperparameters used for each LLM. 
For all models, we set the upper limit for generated tokens on the maximum value allowed by their developers.
We chose to use a temperature of 0.2, i.e., a low value that enables structured, deterministic thinking, while also being higher than zero, allowing a certain degree of exploration.
In the case of o3 and Claude 3.7 Sonnet, we used the temperature value 1, because this was the only temperature value allowed for these models when thinking is enabled.

\subsection{Code Execution Instructions}
\label{appendix:code_execution}

Our code is publicly available on Github\footnote{\url{https://github.com/periklismant/lexicon_neurips}}.
We provide instructions on executing our constrained planning problem generator and our LLM plan verifier on the domains that are present in our repository.
You may add a custom domain by providing a PDDL domain file, an initial state-goal pair generator and NL descriptions of the actions and the atoms of the domain, following the structure of the domains in our repository.

\textbf{Installation.}~You may install \lexicon by following these steps:
\begin{enumerate}
    \item Install conda on an Ubuntu machine.
    \item Clone our repository with Git.
    \item Create a conda environment with the necessary package dependencies installed. To do this, visit the root directory of our repository and run:
    
    \quad\texttt{conda env create $--$name lexiconenv $--$file=environment.yml}

    \item Activate your new conda environment with: \texttt{conda activate lexiconenv}

    \item Make sure that the following packages are installed: [\texttt{anthropic==0.51.0, dotmap==1.3.30, gym==0.26.2, gymnasium==1.0.0, hydra-core==1.3.2, matplotlib==3.7.1, minigrid==3.0.0, numpy==2.2.6, omegaconf==2.3.0, openai==1.81.0, protobuf==6.31.0, pyprover==0.6.2, tqdm==4.67.1, unified\_planning==1.2.0}]

\end{enumerate}

All the instructions that follow require that you have the \texttt{lexiconenv} environment activated.

\textbf{Constrained Planning Problem Generation.}~To generate a constrained planning problem for a specified domain, you may use script \texttt{generate\_benchmark.py}.

This script receives as input:
\begin{itemize}
    \item a domain name ("blocksworld", "babyai", "logistics", "sokoban", or "alfworld"), 
    \item an integer denoting the random seed for generating the first problem in the benchmark,
    \item the number of problems to be generated, and
    \item the number of constraints in each problem.
\end{itemize}

The output of the script is:
\begin{itemize}
    \item  a constrained problem for the domain (in both PDDL and NL), located in folder:
    
    \quad\texttt{domains/\{domain\_name\}/data\_\{constraints\_no\}/\{seed\_no\}}
\end{itemize}

In order to run our problem generator, follow these steps:
\begin{enumerate}
    \item Move into the root directory of our repository.
    \item Construct a directory with the name \texttt{intermediate\_sas}, which is a required folder for $\symk$ to store intermediate computations, with the following command:

    \quad\texttt{mkdir intermediate\_sas}
    
    \item Run command:
    
    \quad\texttt{python3 generate\_benchmark.py \{domain\_name\} \{initial\_seed\} \{problems\_no\} \{constraints\_no\}}
\end{enumerate}

Example executions: 
\begin{itemize}
    \item \texttt{python3 generate\_benchmark.py blocksworld 100 1 2} 
    
    $\rightarrow$ Starting from seed 100, construct a blocksworld problem with 2 constraints. 
    \item \texttt{python3 generate\_benchmark.py logistics 50 3 4}
    
    $\rightarrow$ Starting from seed 50, construct 3 logistics problems with 4 constraints each.
    
\end{itemize}

\textbf{LLM Plan Verification.}~To validate an LLM-generated plan for a constrained planning problem, you may use script \texttt{verify\_plan.py}.

This script receives as input:

\begin{itemize}
    \item a domain name ("blocksworld", "babyai", "logistics", "sokoban", or "alfworld"), 
    \item a folder number (corresponding to the number of constraints in the generated problem), 
    \item a data number (corresponding to the seed used to generate the problem), and 
    \item an llm name ("deepseek", "o3", "gemini-2.5", "claude\_37\_sonnet", "gpt\_5"), where "deepseek" verifies a plan produced by R1.
\end{itemize}

The output of the script is:

\begin{itemize}
    \item an indication on whether the plan stored in
    
    \quad\texttt{domains/\{domain\_name\}/data/data\_\{folder\_no\}/\{data\_no\}/\{llm\}\_plan}
    
    is invalid, suboptimal or optimal.
\end{itemize}

In order to run our LLM plan verifier, follow these steps:
\begin{enumerate}
    \item Move into the root directory of our repository.
    \item Run command:
    
    \quad\texttt{python3 verify\_plan.py \{domain\_name\} \{folder\_no\} \{data\_no\} \{llm\}}
\end{enumerate}

Example executions (on pre-generated, packed LLM plans):
\begin{itemize}
    \item \texttt{python3 verify\_plan.py babyai 1 1 o3} 

    $\rightarrow$ Verifies that the plan in the corresponding directory is optimal.
    \item \texttt{python3 verify\_plan.py babyai 3 1 o3}
    
    $\rightarrow$ Verifies that the plan in the corresponding directory is invalid.
    \item \texttt{python3 verify\_plan.py blocksworld 5 1 o3}
    
    $\rightarrow$ Verifies that the plan in the corresponding directory is suboptimal.
\end{itemize}

\subsection{Experiment Reproducibility Instructions}
\label{appendix:experiment_reproducibility}

Reproducing our experiments requires three main steps:
\begin{enumerate}
    \item Generating benchmarks with problems having an increasing number of constraints for each domain.
    \item Evaluating LLMs on the generated benchmarks.
    \item Verifying the LLM plans produced in the previous step.
\end{enumerate}

Steps 1 and 3 are described in Appendix \ref{appendix:code_execution}.

In order to run LLMs on constrained problems generated by \lexicon, follow these steps:
\begin{enumerate}
    \item Get API keys by OpenAI, Deepseek, Google and Anthropic, and store them in conda environment variables as follows:

    \quad\texttt{conda env config vars set OPENAI\_API\_KEY=yourkey}
    
    \quad\texttt{conda env config vars set DEEPSEEK\_API\_KEY=yourkey}
    
    \quad\texttt{conda env config vars set GEMINI\_API\_KEY=yourkey}
    
    \quad\texttt{conda env config vars set ANTHROPIC\_API\_KEY=yourkey}
    
    You have to deactivate and reactivate your conda environment for the variable changes to take effect.
    In order to use some models, such as o3, you may need to elevate your subscription to a certain tier level.

    \item Open file \texttt{cfg/config.yaml} with a text editor and make the following changes:
    \begin{itemize}
        \item set the value of \texttt{mode} to \texttt{evaluation}.
        \item set the value of \texttt{folder\_no} to the \texttt{constraints\_no} used to generate the problems you want the LLMs to solve.
        \item set the value of list \texttt{evaluation\_data} to the ids of the problems you want to evaluate LLMs on. These problem ids can be found in: 
        
        \quad\texttt{domains/\{domain\_name\}/data/data\_\{folder\_no\}/}
        \item add a new key-value pair ``\texttt{llm:\ evaluation}''.
    \end{itemize}

    \item Evaluate DeepSeek R1, OpenAI o3, Gemini 2.5 Pro, Claude 3.7 Sonnet (with extended thinking) and GPT-5 on the problems selected in the previous step on, e.g., the Blocksworld domain:

    \begin{enumerate}

    \item Create a file named \texttt{run\_blocksworld.py} and add to it the following code:
    \lstset{style=mypython} 
    \begin{lstlisting}
from omegaconf import OmegaConf
from domains.blocksworld.blocksworld import main
if __name__ == "__main__":
    cfg = OmegaConf.load("cfg/config.yaml")
    main(cfg)
    \end{lstlisting}
    
    \item Run the following command:
    
    \quad\texttt{python3 run\_blocksworld.py}

    \end{enumerate}

    You may evaluate these LLMs on a different domain by replacing "blocksworld" with the name of another domain in the above steps.

    \item In order to use different LLMs, open file \texttt{lexicon.py} with a text editor, make the following changes and then go back to the previous step.
    \begin{itemize}
        \item Go to the definition of \texttt{evaluate\_llms} and adjust the elements of list \texttt{model\_names\_and\_strategies}.
    \end{itemize}
\end{enumerate}

\end{document}